\documentclass[10pt,twocolumn,letterpaper]{article}

\usepackage[pagenumbers]{cvpr} % To force page numbers, e.g. for an arXiv version

\usepackage[T1]{fontenc}
\usepackage{multirow}
\usepackage{colortbl}
\definecolor{arcband}{HTML}{CDEEFC}
\usepackage{tcolorbox}
\tcbuselibrary{skins,breakable}
\definecolor{arcblue}{HTML}{3D5FA8}
\definecolor{arcbluebg}{HTML}{F1F5FB}
\newtcolorbox{promptbox}[1]{enhanced, colback=arcbluebg, colframe=arcblue, boxrule=0.6pt, arc=2.5pt,
  left=7pt, right=7pt, top=5pt, bottom=5pt, title={#1}, fonttitle=\bfseries, coltitle=white,
  colbacktitle=arcblue, toptitle=1.5pt, bottomtitle=1.5pt}
\newtcolorbox{codebookbox}[1]{enhanced, breakable, colback=arcbluebg, colframe=arcblue, boxrule=0.6pt, arc=2.5pt,
  left=6pt, right=6pt, top=5pt, bottom=5pt, title={#1}, fonttitle=\bfseries, coltitle=white,
  colbacktitle=arcblue, toptitle=1.5pt, bottomtitle=1.5pt}
\newcommand{\hc}[2]{\cellcolor[HTML]{#1}#2}
\newcommand{\hw}[2]{\cellcolor[HTML]{#1}\textcolor{white}{#2}}
\makeatletter
\newcommand{\arcbandstrut}[2]{\noalign{\xdef\arc@strutht{\the\ht\@arstrutbox}\xdef\arc@strutdp{\the\dp\@arstrutbox}\global\setbox\@arstrutbox\hbox{\vrule height #1 depth #2 width 0pt}}}
\newcommand{\arcrowstrut}{\noalign{\global\setbox\@arstrutbox\hbox{\vrule height \arc@strutht depth \arc@strutdp width 0pt}}}
\makeatother

\newcommand{\arc}{\textbf{AgentRoboCritic}\xspace}

\newcommand{\cat}[1]{\textsc{#1}\xspace}
\newcommand{\Achieved}{\cat{achieved}}
\newcommand{\Failed}{\cat{failed}}
\newcommand{\Partial}{\cat{partial}}
\newcommand{\Unverifiable}{\cat{unverifiable}}
\newcommand{\NotApplicable}{\cat{not applicable}}
\newcommand{\Grasp}{\cat{Grasp}}
\newcommand{\Alignment}{\cat{Alignment}}
\newcommand{\Interference}{\cat{Interference}}
\newcommand{\Effect}{\cat{Effect}}
\newcommand{\IncMotion}{\cat{Incomplete motion}}
\newcommand{\Choice}{\cat{Choice}}
\newcommand{\Unresolved}{\cat{Unresolved}}
\newcommand{\Closure}{\cat{Closure}}
\newcommand{\Handover}{\cat{Handover}}
\newcommand{\Lift}{\cat{Lift}}
\newcommand{\Carry}{\cat{Carry}}
\newcommand{\Release}{\cat{Release}}

\definecolor{citecolor}{RGB}{0,100,80}
\AtEndPreamble{%
  \crefname{section}{Section}{Sections}%
  \crefname{table}{Table}{Tables}%
  \crefname{figure}{Figure}{Figures}%
  \crefname{equation}{Equation}{Equations}%
}

\definecolor{cvprblue}{rgb}{0.21,0.49,0.74}
\usepackage[pagebackref,breaklinks,colorlinks,allcolors=cvprblue,hypertexnames=false]{hyperref}
\usepackage{etoc}

\def\confName{CVPR}
\def\confYear{2027}

\title{AgentRoboCritic: Understanding the Capability of Frontier Agents on Direct Robot Control}

\newcommand{\arcauthors}{%
Yiyang Du$^{1\dagger}$ \quad Yuxuan Kuang$^{1*}$ \quad Qitao Zhao$^{1*}$ \quad Yixuan Wang$^{2*}$ \quad Zhanqiu Guo$^{1}$\\
Yi Xu$^{3}$ \quad Xin Li$^{3}$ \quad Wenbin He$^{3}$ \quad Liu Ren$^{3}$ \quad Chenyan Xiong$^{1\dagger}$\\[0.4em]
{\normalsize $^{1}$Carnegie Mellon University \quad $^{2}$University of Illinois at Urbana-Champaign}\\
{\normalsize $^{3}$Bosch Research North America \& Bosch Center for Artificial Intelligence (BCAI)}}
\newcommand{\arcauthornotes}{%
{\let\thefootnote\relax\footnotetext{\raggedright\hspace*{-1.8em}$^*$Equal contribution.\\$^\dagger$Corresponding authors: \texttt{\{yiyangd,cx\}@cs.cmu.edu}\\$^{1}$Project page: \href{https://adu2021.github.io/blog/AgentRoboCritic/}{\texttt{adu2021.github.io/blog/AgentRoboCritic}}}}}
\author{\arcauthors}

\begin{document}
\maketitle
\arcauthornotes
\begin{abstract}
Frontier multimodal language agents such as GPT-6 Astra have demonstrated strong capabilities in controlling robots directly via low-level numerical actions without high-level primitives. However, existing benchmarks report only episode-level success rates, which show neither how these agents succeed nor where they fail. In this paper, we present a systematic action-level analysis of frontier agents on direct robot control across 12 manipulation benchmarks. We introduce \textbf{AgentRoboCritic}, an intent-grounded critique framework in which the policy agent states a local intent alongside each action, and a critic evaluates the action's outcome and failure mode against this intent. Our action-level analysis demonstrates the promising capabilities of frontier agents on direct robot control, including closed-loop recovery from local execution failures, reliable semantic grounding across diverse scenes, and strong zero-shot manipulation without task-specific adaptation. It also identifies their current limitations, including task-relevant physical execution bottlenecks such as grasping and alignment, and uneven recovery across failure modes. Finally, we validate our critic on a 400-action human-annotated benchmark and release a dataset of 12,844 annotated actions alongside a fine-tuned open-source critic, Muse-Glimmer-30B, whose failure-mode accuracy rises from 32.58\% to 74.24\%, enabling low-cost action-level analysis.$^{1}$
\end{abstract}
% Teaser (Figure 1): a figure* declared on page 1, so it is placed at the top of page 2
% Full-width teaser, declared on page 1 so that it lands at the top of page 2.
\begin{figure*}[t]
\centering
\includegraphics[width=\textwidth]{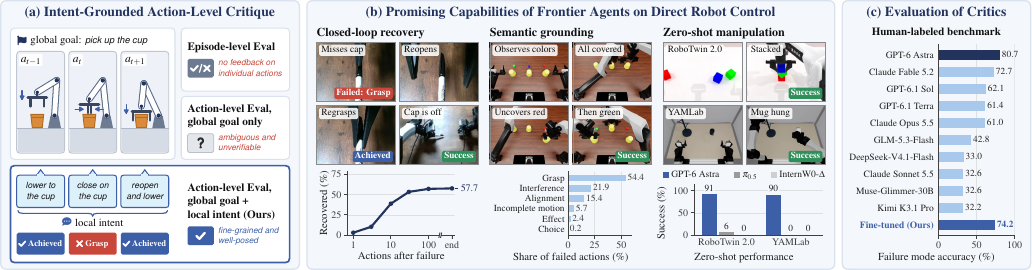}
\caption{\textbf{Overview of \arc.} (a) Local intent makes action-level critique fine-grained and well-posed. (b) Capabilities of frontier agents on direct robot control. (c) Evaluation of action-level critics on the human-labeled benchmark.}
\label{fig:teaser}
\end{figure*}

\etocdepthtag.toc{mtmain}
\section{Introduction}
\label{sec:intro}

Recent advances in robot foundation models follow two main paradigms: Vision-Language-Action models (VLAs), which adapt Vision-Language Models (VLMs) to output low-level robot actions, and World-Action Models (WAMs), which build upon video-generation world models. Despite building upon existing foundation models, both families still suffer from brittle out-of-distribution generalization~\cite{fei2025liberoplus,zhang2026liberomax,wang2025vlatest} and limited long-horizon memory~\cite{dai2026robomme,shi2026memoryvla}. Moreover, deploying VLAs and WAMs to new downstream tasks typically requires task-specific fine-tuning~\cite{kim2024openvla,black2024pi0,black2025pi05,ye2026worldactionmodelszeroshot}, limiting their zero-shot adaptation.

Beyond these two paradigms, frontier multimodal language agents such as GPT-6 Astra~\cite{openai2026gpt6astra} offer a promising alternative: they control robots directly via low-level numerical actions without relying on high-level primitives, show strong zero-shot generalization and working memory in robotic manipulation~\cite{astra2026early,robomme2026frontier,bitterlesson2026,robocoliseum2026}, and outperform specialized VLAs and WAMs~\cite{astra2026early,robomme2026frontier,tactalent}. These results demonstrate the vast potential of leveraging frontier language model agents directly as generalist robot policies.

Existing evaluations assess these agents at the \textit{episode level}, with binary success rates or aggregate task scores that show neither how an agent succeeds nor where it fails. In this paper, we provide a systematic, fine-grained investigation into the capabilities and limitations of frontier agents on direct robot control by analyzing their behavior at the \textit{action level}. Evaluating an individual action can be ambiguous because robot manipulation is inherently multimodal~\cite{florence2022ibc,shafiullah2022bet,chi2023diffusion}: since multiple distinct trajectories can validly advance the same task, whether an action succeeds depends on its intended sub-goal. We resolve this ambiguity by letting the policy agent state a local intent alongside each action, establishing a verifiable local contract for every action (Figure~\ref{fig:teaser}(a)). Conditioned on the global task goal, this local intent, and multi-view observations, our \textbf{\arc} framework determines whether each executed action succeeded and assigns a specific failure mode to every failed action. Using GPT-6 Astra as the critic, we annotate episodes across 12 robot manipulation benchmarks for further analysis.

Our analysis characterizes both the promising capabilities and the current limitations of frontier agents on direct robot control. On the capability side, frontier agents show three major strengths (Figure~\ref{fig:teaser}(b)). First, they recover from local execution failures in closed loop, unlike imitation-trained policies whose errors compound under covariate shift~\cite{ross2011dagger}, and half of their successful episodes contain at least one failed action that the agent later corrects. Second, they show reliable semantic grounding across diverse scenes, executing actions that follow the target object, category, and sequential order specified in the local intent. Third, they achieve strong zero-shot manipulation without task-specific adaptation. On the limitation side, frontier agents face task-relevant physical execution bottlenecks, led by object acquisition and retention (\Grasp) in general manipulation, spatial mating (\Alignment) in tight-tolerance visuo-tactile tasks, and unintended collisions (\Interference) in cluttered scenes. Recovery is also uneven across failure modes: \Alignment{} failures are recovered less often than \Grasp{} failures.

To validate the reliability of our automated annotations, we construct a 400-action human-annotated evaluation benchmark and demonstrate that GPT-6 Astra aligns closely with human judgments, while ablating local intent causes a sharp collapse in failure-mode attribution. Finally, to enable low-cost action-level critique for direct robot control, we curate an open-source dataset of 12,844 annotated actions and fine-tune an open-source critic VLM, Muse-Glimmer-30B, which approaches GPT-6 Astra on both action outcome and failure-mode accuracy (Figure~\ref{fig:teaser}(c)).

Our main contributions are as follows:
\begin{itemize}
  \setlength{\itemsep}{2pt}
  \setlength{\parskip}{0pt}
  \item \textbf{Intent-Grounded Action Critique Framework:} We propose \textbf{\arc}, a framework that evaluates frontier language model agents on direct robot control at the action level by conditioning the critique on both global task objectives and the policy's local intent.
  \item \textbf{Action-Level Analysis of Frontier Agents on Direct Robot Control:} We go beyond episode-level success rates to establish frontier agents as a promising path toward generalist robot policies and to identify the physical execution bottlenecks that remain.
  \item \textbf{Open-Source Critique Dataset and Critic VLM:} We curate and release an action-level critique dataset alongside a 400-action human-annotated test set, as well as a fine-tuned Muse-Glimmer-30B critic VLM that enables accurate, low-cost failure analysis.
\end{itemize}

\section{Related Work}
\label{sec:related}

\paragraph{Frontier Agents as Robot Policies.}
Language models have been connected to robots through planning over pretrained skills~\cite{ahn2022saycan}.
Code-as-policy approaches generate programs that call perception and control APIs~\cite{liang2023cap,tsui2026faea,urai2026codeaspolicy}.
Both lines of work delegate execution to skills and control primitives.
Recent studies investigate frontier multimodal language agents that generate numerical actions directly from observations~\cite{astra2026early,robocoliseum2026,bitterlesson2026,robomme2026frontier}.
We explore the potential of frontier agents in direct robot control, which reflects their intrinsic embodied abilities disentangled from those of the primitives.

\paragraph{Evaluation on Frontier Agents in Direct Robot Control.}
A growing set of studies measures frontier agents acting as the robot policy.
Reports on GPT-6 Astra score it against public policies on all 42 RoboDojo tasks~\cite{chen2026robodojo,astra2026early}, on the memory-dependent tasks of RoboMME~\cite{dai2026robomme,robomme2026frontier}, on the simulated RoboLab-120 pick-and-place suite~\cite{astranomy}, on real bimanual desk tasks~\cite{stationerybench}, and on visuo-tactile UniVTAC and OpenViTac tasks~\cite{tactalent}.
Cross-model studies run several frontier models through one shared harness, across the four RoboColiseum tracks~\cite{robocoliseum2026} and on physical arms for six RoboDojo tasks, where capability is tracked over time~\cite{bitterlesson2026}.
GPT-GEAR probes depth, in-context learning and tactile use, then scores GPT-6 Astra on five established benchmarks and a new geometry-constrained suite~\cite{zhou2026gptgear}.
The open harness Embodied First Exam~\cite{efe} exposes hundreds of simulated tasks to coding agents and verifies outcomes independently of what the agent reports.
These evaluations assess behavior at the episode level via binary success rates or aggregate progress scores. \arc instead judges each action against the local intent, providing fine-grained and well-posed feedback.

% Figure 2 (method overview, Section 3.2) is declared here so that it lands at the top of page 3.
\begin{figure*}[t]
\centering
\includegraphics[width=\textwidth]{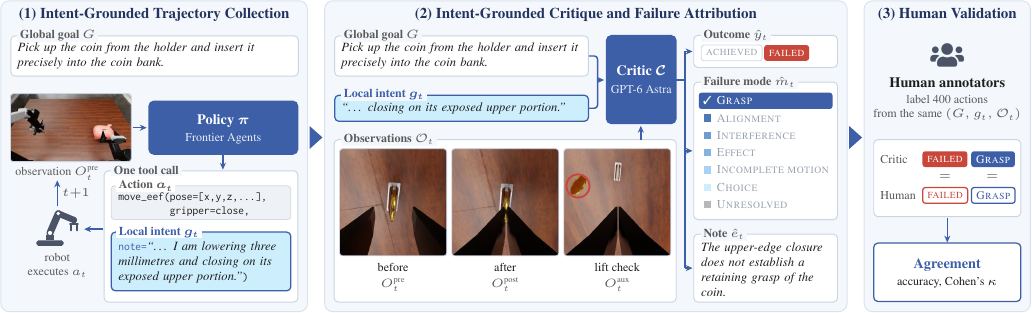}
\caption{\textbf{Intent-grounded action-level critique.} (1) The policy agent states a local intent $g_t$ alongside every action $a_t$, in the same tool call. (2) The critic receives the global goal $G$, the local intent $g_t$ and the observations $\mathcal{O}_t$, and returns the outcome, the failure mode and a note. (3) Human annotators label the same actions to validate the reliability of critics.
% The example is a real action from RoboDojo: the gripper closes on the upper edge of the coin, which drops in the lift check (red circle), so the critic labels the action \Failed with failure mode \Grasp.
}
\label{fig:method_overview}
\end{figure*}

\paragraph{Robot Failure Detection and Reasoning.}
REFLECT~\cite{liu2023reflect} summarizes robot experience for explanation and correction.
AHA~\cite{duan2025aha} trains a critic using procedurally perturbed demonstrations; RoboFAC~\cite{robofac2025} combines failure data, analysis and correction; ViFailback~\cite{zeng2026vifailback} studies real VLA failures with textual and visual corrective guidance.
FailBench~\cite{failbench2026} evaluates task-success judgments across sources.
\arc focuses on failures arising naturally in frontier-agent trajectories and grounds action-level critique in the agent's local intent.

\section{Method}
\label{sec:method}

To systematically understand the capabilities and current limitations of frontier agents in physical manipulation, we first formulate the paradigm of \emph{direct robot control} (Section~\ref{sec:direct_control}) and then introduce \arc, our intent-grounded failure attribution framework that diagnoses agent execution at the granularity of individual actions (Section~\ref{sec:critic_method}).

\subsection{Frontier Agents on Direct Robot Control}
\label{sec:direct_control}

We define \emph{direct robot control} as a setting in which a frontier multimodal language agent directly generates numerical robot control commands from a global task instruction, visual observations, and interaction history. Consider a manipulation episode governed by a task instruction $G$ (e.g., \emph{``open the box, take out the eraser, and close the lid''}) over discrete decision turns $t \in \{1, \dots, T\}$. At each turn $t$, given $G$, the current visual observation $O_t^{\text{pre}}$, and the interaction history $h_t$, the frontier agent policy $\pi$ outputs a control command $a_t$:
\begin{equation}
  a_t \sim \pi\!\left(\cdot \mid G,\, O_t^{\text{pre}},\, h_t\right),
  \label{eq:direct_control}
\end{equation}
where $a_t$ specifies low-level numerical targets such as 6-DoF end-effector poses, joint configurations, gripper openings, or primitive motion trajectories. The robot executes $a_t$ in the physical environment and returns the updated observation $O_t^{\text{post}}$, forming a closed-loop control cycle.

Unlike specialized VLAs or WAMs, which require additional action heads or continuous policy decoders atop pretrained vision-language or video-generation backbones, frontier agents express these numerical control actions as structured text through their native language-generation interface.
% Specifically, the agent invokes structured tool calls carrying the numerical control parameters, which are parsed and dispatched directly to the robot's low-level control stack for execution.
To assess \textbf{zero-shot} capability on robot manipulation, we evaluate frontier agents without task-specific fine-tuning or in-context demonstrations, and compare their performance against VLAs and WAMs under the same zero-shot evaluation protocol.

\subsection{Intent-Grounded Action-Level Critic}
\label{sec:critic_method}

% Figure 3 (failure modes) is declared here so that it lands at the top of page 4.
\begin{figure*}[t]
\centering
\includegraphics[width=\textwidth]{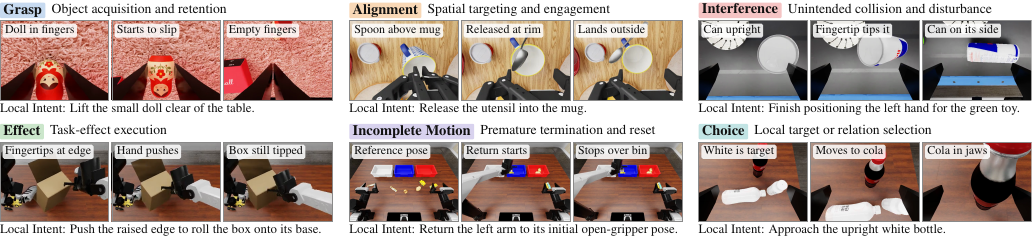}
\caption{\textbf{Action-level failure modes ($\mathcal{M}$) in \arc.} Each of the six modes shown is illustrated by one failed action from our annotated episodes. Apart from the six modes, \Unresolved marks failures with occluded or ambiguous evidence.}
\label{fig:failure_modes}
\end{figure*}

While direct robot control allows frontier agents to solve complex manipulation tasks zero-shot, standard episode-level success rates obscure how individual actions succeed or fail during closed-loop execution. Figure~\ref{fig:method_overview} illustrates the overall pipeline of \arc. Below, we formulate intent-grounded action critique, then describe trajectory collection, failure attribution and the human-annotated benchmark.

\noindent\textbf{Task Formulation.}
Critiquing an isolated robotic action $a_t$ solely against the global task instruction $G$ is ill-posed because robotic manipulation is inherently multimodal, allowing many distinct trajectories to validly advance the same overarching goal. To establish a concrete, verifiable local contract for step $t$, we extend the direct control formulation in Equation~\eqref{eq:direct_control} so that the policy $\pi$ jointly generates the low-level control command $a_t$ and a natural-language local intent $g_t$ specifying the immediate sub-goal of that action, yielding an intent-grounded trajectory $\tau$:
\begin{equation}
  \begin{gathered}
    \left(a_t,\, \textcolor{blue}{g_t}\right) \sim \pi\!\left(\cdot \mid G,\, O_t^{\text{pre}},\, h_t\right), \\
    \tau = \left\{\left(O_t^{\text{pre}},\, a_t,\, \textcolor{blue}{g_t},\, O_t^{\text{post}}\right)\right\}_{t=1}^{T}.
  \end{gathered}
  \label{eq:trajectory_generation}
\end{equation}
Executing $a_t$ induces a physical state transition captured by a bounded visual observation window $\mathcal{O}_t = \left(O_t^{\text{pre}}, O_t^{\text{post}}, O_t^{\text{aux}}\right)$, where $O_t^{\text{pre}}$ and $O_t^{\text{post}}$ denote the multi-view camera observations immediately preceding and following action execution, and $O_t^{\text{aux}}$ denotes optional short-horizon follow-up frames (such as the immediate subsequent lift after a gripper closure) required to verify physical object capture and retention.

Conditioned on the self-generated local intent $g_t$, action-level evaluation becomes well-posed. We formulate the intent-grounded action-level critic as a mapping $\mathcal{C}$:
\begin{equation}
  (\hat{y}_t,\, \hat{m}_t,\, \hat{e}_t) = \mathcal{C}\!\left(G,\, g_t,\, \mathcal{O}_t\right),
  \label{eq:critic_formulation}
\end{equation}
where $\hat{y}_t \in \mathcal{Y}$ is the action-level outcome label, $\hat{m}_t \in \mathcal{M} \cup \{\varnothing\}$ is the attributed failure mode when the action fails ($\hat{m}_t = \varnothing$ if successful), and $\hat{e}_t$ is a brief natural-language diagnostic note explaining the physical evidence supporting $(\hat{y}_t, \hat{m}_t)$. In our general annotation protocol, $\mathcal{Y} = \{\text{Achieved}, \text{Failed}, \text{Partial}, \text{Unverifiable}\}$ to account for intentionally incremental movements (\text{Partial}) and occluded or truncated recordings (\text{Unverifiable}), which reduces to binary success classification on completed actions.

\noindent\textbf{Intent-Grounded Trajectory Collection.}
Standard robot evaluation rollouts typically log only low-level states and actions without action-level local intents, and retrofitting end-to-end VLAs or WAMs to articulate reliable sub-goals at every control step is non-trivial~\cite{zawalski2024ecot,sun2026revisitingecot}. Frontier agents, in contrast, already act through structured tool calls. We extend each tool call with a mandatory \texttt{note} parameter in which the agent states the local intent $g_t$ of each commanded action $a_t$ (Equation~\eqref{eq:trajectory_generation}). Any tool call that omits this local intent parameter is automatically rejected by the execution harness, prompting the agent to reissue a properly grounded command. Beyond live rollouts, we also curate released evaluation trajectories from open benchmarks where action-level decision notes and synchronized visual observations were natively logged~\cite{chen2026robodojo,stationerybench,tactalent}.

\noindent\textbf{Intent-Grounded Critique and Failure Attribution.}
Given each action record $(G, g_t, \mathcal{O}_t)$, we employ GPT-6 Astra as an automated action-level critic to generate structured annotations $(\hat{y}_t, \hat{m}_t, \hat{e}_t)$. To ensure physical rigor and prevent hallucinated causes, our annotation protocol enforces three core design principles:
(1)~\emph{Contact-aware temporal windows:} Evaluating whether an action achieved its local intent requires matching the observation window to the physical semantics of $g_t$. While reaching and alignment steps can be judged from action-boundary frames $(O_t^{\text{pre}}, O_t^{\text{post}})$, gripper-closing actions require evidence of stable capture and retention, which we provide via immediate post-grasp lift frames $O_t^{\text{aux}}$.
(2)~\emph{Decoupling local validity from episode outcomes:} An action is judged strictly against its local intent $g_t$ and visible physical transition. An early failed grasp that the agent subsequently recovers from is still labeled as an action-level failure. Conversely, a deliberate diagnostic test (e.g., slightly loosening the gripper to test stability) or an intentionally small intermediate adjustment is labeled \emph{Achieved} when it fulfills $g_t$.
(3)~\emph{Evidence-bounded failure attribution:} For every action judged as \emph{Failed}, the critic classifies the failure into one of the seven failure modes illustrated in Figure~\ref{fig:failure_modes} (\Grasp, \Alignment, \Interference, \Effect, \IncMotion, \Choice, or \Unresolved) and writes a concise diagnostic note $\hat{e}_t$ specifying the active arm, target object, contact geometry, and observed failure timing. More details are provided in Appendix~\ref{sec:supp_taxonomy}.

% Figure for Section 4, placed here so that the two-column float is read on page 4 and lands on top of page 5.
\begin{figure*}[t]
\centering
\includegraphics[width=\textwidth]{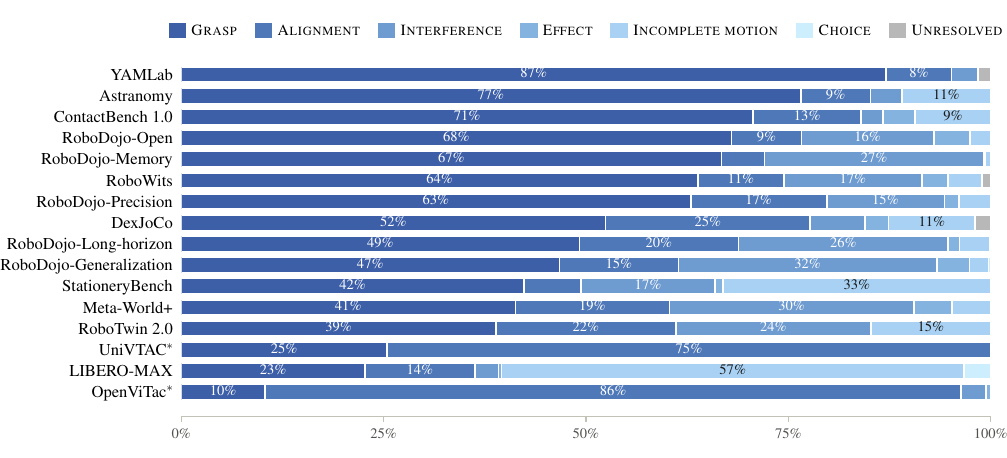}
\caption{\textbf{Action-level failure modes per benchmark.} Each bar shows the distribution of failure modes among the failed actions of one benchmark. $^\ast$ denotes the vision-only settings of visuo-tactile benchmarks. RoboDojo is shown by its 5 evaluation dimensions separately.}
\label{fig:mechanisms}
\end{figure*}

\noindent\textbf{Human-Annotated Benchmark.}
We build a human-annotated benchmark of 400 actions across the manipulation benchmarks, comprising 34\% successful and 66\% failed actions. Annotators receive the same input $(G, g_t, \mathcal{O}_t)$ as the critic and independently label each action with its outcome and, if failed, its failure mode. Section~\ref{sec:analysis} evaluates all critics against these labels.

\section{Experiments}
\label{sec:exp}

We design our experimental evaluation to systematically characterize both the capabilities and the current limitations of frontier agents in direct robot control.
We assemble an evaluation suite of GPT-6 Astra rollouts in direct robot control across 12 manipulation benchmarks:
\textit{RoboDojo}~\cite{chen2026robodojo},
\textit{StationeryBench}~\cite{stationerybench},
\textit{Astranomy}~\cite{astranomy},
\textit{UniVTAC}~\cite{chen2026univtac},
\textit{OpenViTac}~\cite{wu2026openvitac},
\textit{ContactBench 1.0}~\cite{xiang2026contactbench},
\textit{LIBERO-MAX}~\cite{zhang2026liberomax},
\textit{Meta-World+}~\cite{mclean2025metaworldplus},
\textit{DexJoCo}~\cite{wang2026dexjoco},
\textit{RoboWits}~\cite{lin2026robowits},
\textit{RoboTwin 2.0}~\cite{chen2025robotwin2}, and
\textit{YAMLab}~\cite{yamlab}. We then analyze this collection with our \arc framework. We verify the reliability of its critic against human annotations in Section~\ref{sec:analysis}.

\noindent\textbf{Implementation Details.} For seven benchmarks (\textit{LIBERO-MAX}, \textit{Meta-World+}, \textit{DexJoCo}, \textit{RoboWits}, \textit{RoboTwin 2.0}, \textit{YAMLab}, and \textit{ContactBench 1.0}), we collect new GPT-6 Astra episodes, requiring a local intent alongside every action and recording camera observations at action boundaries. For the remaining five benchmarks, we ingest released GPT-6 Astra evaluation traces through unified alignment adapters that pair the intent notes with pre- and post-action frames. We compare frontier agents against representative VLAs and WAMs, including $\pi_{0.5}$~\cite{black2025pi05} and InternW0-$\Delta$~\cite{miao2026internw0delta}, under the same zero-shot setting without task-specific fine-tuning. More implementation details are provided in Appendix~\ref{sec:supp_implementation}.

\subsection{Capabilities and Promise of Frontier Agents}
\label{sec:exp_capabilities}

We first examine what frontier agents already do well in direct robot control and why this paradigm holds strong promise for generalist manipulation. Figure~\ref{fig:mechanisms} summarizes the action-level failure distributions across all evaluated benchmarks, revealing that while failure patterns vary with task structure and concentrate primarily in low-level physical grasping (\Grasp), semantic selection errors (\Choice) remain remarkably rare. Building on our action-level critique and zero-shot policy evaluations, we highlight three core strengths of frontier agents: closed-loop recovery from local execution failures, reliable semantic grounding across diverse scenes, and strong zero-shot manipulation without task-specific adaptation.

\begin{figure*}[t]
\centering
\includegraphics[width=\textwidth]{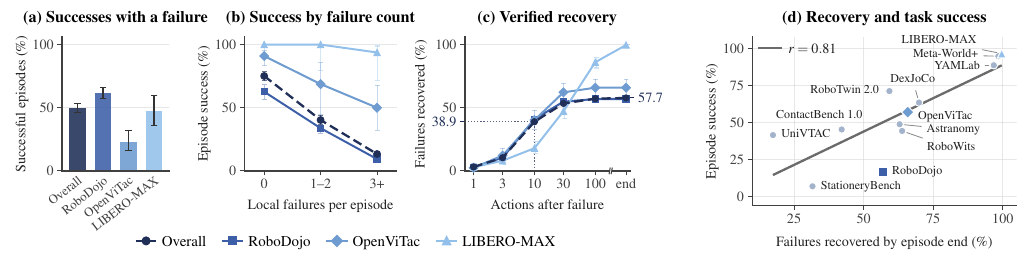}
\caption{\textbf{Closed-loop recovery from local execution failures.} Overall combines all 12 benchmarks.
(a) Share of successful episodes that contain at least one failed action.
(b) Episode success rate conditioned on the number of local failures within an episode.
(c) Share of failed actions with a verified recovery within a given number of later actions, up to the end of the episode.
(d) Verified recovery rate against the official success rate of episodes with a local failure.}
\label{fig:recoverability}
\end{figure*}

\noindent\textbf{Closed-Loop Recovery from Local Execution Failures.}
A classic bottleneck of imitation-trained robot policies is covariate shift~\cite{ross2011dagger}, where a single execution error pushes the robot outside the demonstration distribution and subsequent errors compound until the episode fails. Our action-level annotations reveal that frontier agents in direct control behave differently, recovering from local failures and still completing the task. As shown in Figure~\ref{fig:recoverability}(a), 49.8\% of all successful episodes contain at least one local failure, showing that half of the successes are reached through recovery. Episodes with one or two local failures still succeed in 40.3\% of cases, compared to 75.1\% for failure-free episodes (Figure~\ref{fig:recoverability}(b)), indicating that frontier agents complete a substantial share of tasks despite local failures. These results show that frontier agents tolerate and correct their own execution errors.

We further quantify the failure recovery dynamics of frontier agents on direct robot control. We analyze failed actions across the 12 benchmarks and count a failure as recovered when a subsequent action in the same episode verifiably resolves it, such as a regrasp after which the object rises with the gripper. In Figure~\ref{fig:recoverability}(c), 38.9\% of the analyzed failures are recovered within 10 actions and 57.7\% by the end of the episode, showing that frontier agents resolve more than half of their own failures. Figure~\ref{fig:recoverability}(d) further shows that benchmarks with a higher recovery rate also reach a higher official success rate among the episodes that contain a local failure, with a Pearson correlation of 0.81, indicating that failure recovery is closely associated with task success in direct robot control. This demonstrates that frontier agents can act as closed-loop policies with promising failure recovery ability (Appendix~\ref{sec:supp_recovery}).

\noindent\textbf{Reliable Semantic Grounding Across Diverse Scenes.}
Specialized VLAs and WAMs remain vulnerable to visual perturbations, camera shifts, and spatial rearrangements, often suffering steep performance drops on robustness suites such as LIBERO-Plus, LIBERO-MAX, and VLATest~\cite{fei2025liberoplus,zhang2026liberomax,wang2025vlatest}. Our action-level critique highlights a clear contrast for frontier agents. A failed action is labeled \Choice when its execution contradicts the object, order or category stated in the local intent. As shown in Figure~\ref{fig:mechanisms}, \Choice accounts for 0.2\% of all failed actions across the 12 benchmarks and for at most 3.3\% on any single benchmark. Even when confronted with novel object instances and visual distractors, frontier agents act on the object, order and category stated in their local intent, demonstrating reliable semantic grounding in direct robot control.

\begin{figure*}[t]
\centering
\includegraphics[width=\textwidth]{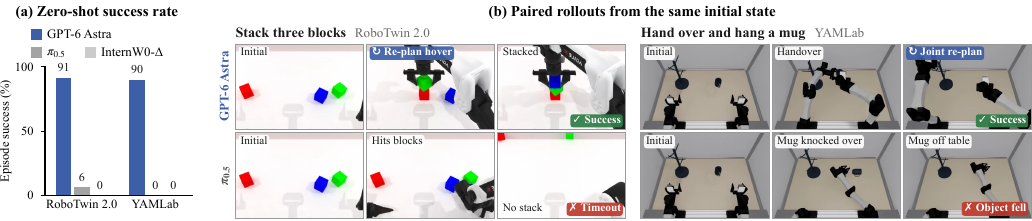}
\caption{\textbf{Zero-shot direct robot control comparison against VLAs and WAMs.} (a) Head-to-head zero-shot episode success rates comparing a frontier agent (GPT-6 Astra) against representative VLA and WAM ($\pi_{0.5}$ and InternW0-$\Delta$) without task-specific fine-tuning. (b) Qualitative rollout trajectories illustrating successful zero-shot task completion and closed-loop adjustment by the frontier agent alongside failure modes of the VLA baseline.}
\label{fig:zeroshot_comparison}
\end{figure*}

\noindent\textbf{Strong Zero-Shot Manipulation.}
Achieving zero-shot generalization to unseen manipulation tasks and environments without in-domain demonstrations or task-specific adaptation has been a long-standing goal in robot learning~\cite{openx2024,brohan2023rt2,octo2024,black2024pi0}. However, existing VLAs and WAMs, including recent generalist policies such as $\pi_{0.5}$~\cite{black2025pi05}, typically struggle when deployed zero-shot outside their training distribution and require task-specific fine-tuning to achieve reliable control. As shown in Figure~\ref{fig:zeroshot_comparison}, frontier multimodal language agents such as GPT-6 Astra achieve strong zero-shot success rates across diverse manipulation benchmarks directly out of the box, substantially outperforming unadapted VLA and WAM baselines. Moreover, frontier agents achieve this manipulation capability while preserving their general-purpose language reasoning and multimodal understanding within a single unified model. By contrast, training or adapting VLAs on domain-specific robot trajectories frequently degrades the pretrained VLM backbone's general visuo-linguistic reasoning and conversational capabilities~\cite{zhou2025chatvla,zhou2025chatvla2,yang2025instructvla}.

\subsection{Current Limitations of Frontier Agents}
\label{sec:exp_limitations}

\noindent\textbf{Task-Relevant Physical Execution Bottlenecks.}
Despite these capabilities, frontier agents still face task-relevant physical execution bottlenecks. As shown in Figure~\ref{fig:mechanisms}, action-level failure modes are concentrated in low-level physical interaction and shift systematically with task geometry. In general tabletop and bimanual manipulation, object acquisition and retention (\Grasp) is the primary bottleneck and the most frequent failure mode on nine of the ten non-tactile benchmarks. When the task setting moves to tight-tolerance assembly and insertion on the vision-only visuo-tactile benchmarks (\emph{UniVTAC} and \emph{OpenViTac}), the primary bottleneck shifts sharply from \Grasp{} to \Alignment{}, as the agent reliably grasps the connector or block and approaches the target receptacle, but struggles to resolve millimeter-scale mating tolerances and angular alignment from vision alone. Meanwhile, benchmarks with dense object clutter or multi-stage rearrangement (such as \textit{RoboDojo-Generalization}, \textit{RoboDojo-Memory}, \textit{RoboDojo-Long-horizon}, and \textit{Meta-World+}) exhibit a marked rise in \Interference{} due to unintended collisions during transit, whereas LIBERO-MAX and StationeryBench exhibit a high share of \IncMotion{} from prematurely terminated reaches or incomplete releases.

\begin{figure}[t]
\centering
\includegraphics[width=\columnwidth]{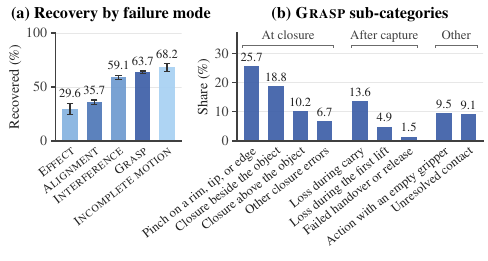}
\caption{\textbf{Recovery by failure mode and \Grasp{} sub-categories.} (a) Share of failed actions with a verified recovery by the end of the episode. (b) Share of each sub-category among all \Grasp{} failures across the 12 benchmarks.}
\label{fig:failure_analysis}
\end{figure}

\noindent\textbf{Uneven Recovery Across Failure Modes.}
Frontier agents recover from most \IncMotion{}, \Grasp{} and \Interference{} failures by the end of the episode, but from only one in three \Alignment{} and \Effect{} failures (Figure~\ref{fig:failure_analysis}(a)), and additional actions do not close this gap (Appendix~\ref{sec:supp_recovery}). In these two modes the action reaches its target and fails on the final offset or on the physical effect of the contact. This shows that closed-loop recovery corrects failures of object acquisition and motion but leaves most precision-limited failures unresolved, indicating that  spatial and contact precision remains essential for tight-tolerance manipulation.

\subsection{Fine-Grained Analysis of Grasp Failures}
\label{sec:grasp_analysis}

\Grasp{} is the most frequent failure mode in direct robot control (Figure~\ref{fig:mechanisms}). To locate the point at which a grasp fails, the critic assigns every \Grasp{} failure a sub-category that records the contact error or the moment of loss. Figure~\ref{fig:failure_analysis}(b) reports their shares across all 12 benchmarks.

\noindent\textbf{Grasps Fail at Closure.}
Three in five \Grasp{} failures occur as the gripper closes, before the object is held, and only one in five occurs after capture. At closure, the gripper most frequently pinches only a rim, tip, or edge of the object, or closes beside or above it. In each of these cases the agent has selected the correct object and brought the gripper to it, and the grasp fails due to a small offset in depth or lateral position. This demonstrates that the grasping bottleneck of frontier agents is the fine 3D localization of the gripper relative to the object. Closure is also the leading stage on every benchmark with more than 100 \Grasp{} failures (Appendix~\ref{sec:supp_results}), indicating that higher spatial precision is the most direct path to more reliable direct robot control.

\noindent\textbf{Failures After Capture.}
Once an object is captured, handover and release account for only 1.5\% of \Grasp{} failures and most losses occur during carrying, showing that the remaining risk lies in transport. A further 9.5\% of \Grasp{} failures are actions issued with an already empty gripper, in which the agent continues to lift or carry an object it has lost. This shows that frontier agents can miss a lost object in their observations, and indicates that an action-level critic can serve as an online monitor that reports the loss at the action where it occurs.

\section{Evaluation of Action-Level Critics}
\label{sec:analysis}
\label{sec:critic_reliability}

\begin{table*}[t]
\centering
\caption{\textbf{Action-level critic evaluation on the human-annotated test set.} Outcome is scored on all actions and Failure Mode on the failed actions. Overall Accuracy requires both to be correct. Inference costs on the test set are reported in US dollars. Bold and underlined values mark the best and second-best results.}
\vspace{4pt}
\footnotesize
\edef\arcsaveabove{\the\aboverulesep}
\edef\arcsavebelow{\the\belowrulesep}
\begin{tabular}{l | c c c c c c}
\toprule
\multirow{2}{*}[-0.5ex]{\textbf{Critic Model}}
& \multicolumn{2}{c}{\textbf{Outcome}\rule[-1.3ex]{0pt}{0pt}}
& \multicolumn{2}{c}{\textbf{Failure Mode}}
& \multirow{2}{*}[-0.5ex]{\textbf{Overall Accuracy (\%)}}
& \multirow{2}{*}[-0.5ex]{\textbf{Cost (\$)}} \\
\noalign{\global\aboverulesep=0pt\global\belowrulesep=0pt}
\cmidrule(lr){2-3} \cmidrule(lr){4-5}
\noalign{\global\aboverulesep=\arcsaveabove\global\belowrulesep=\arcsavebelow}
\rule{0pt}{2.4ex}
& \textbf{Acc. (\%)} & \textbf{$\kappa$}
& \textbf{Acc. (\%)} & \textbf{$\kappa$} & & \\
\noalign{\global\belowrulesep=0pt}
\midrule
\noalign{\global\belowrulesep=\arcsavebelow\global\aboverulesep=0pt}
\arcbandstrut{7.2pt}{2.5pt}
\rowcolor{arcband}\multicolumn{7}{c}{\textbf{\textit{Proprietary Models}}} \\
\arcrowstrut
\midrule
\noalign{\global\aboverulesep=\arcsaveabove}
GPT-6 Astra & \textbf{79.25} & \underline{0.497} & \textbf{80.68} & \textbf{0.600} & \textbf{71.25} & 26.06 \\
\quad w/o local intent $g_t$ & 25.25 & 0.102 & 25.00 & 0.078 & 24.50 & 25.18 \\
GPT-6.1 Sol & 74.50 & 0.487 & 62.12 & 0.314 & 69.75 & 5.00 \\
GPT-6.1 Terra & 74.50 & 0.470 & 61.36 & 0.303 & 66.75 & 2.36 \\
Claude Fable 5.2 & \underline{78.25} & \textbf{0.523} & 72.73 & 0.468 & \underline{71.00} & 26.52 \\
Claude Opus 5.5 & 72.50 & 0.434 & 60.98 & 0.323 & 66.00 & 12.76 \\
Claude Sonnet 5.5 & 56.75 & 0.223 & 32.58 & 0.120 & 50.75 & 5.51 \\
\noalign{\global\belowrulesep=0pt}
\midrule
\noalign{\global\belowrulesep=\arcsavebelow\global\aboverulesep=0pt}
\arcbandstrut{7.2pt}{2.5pt}
\rowcolor{arcband}\multicolumn{7}{c}{\textbf{\textit{Open-Source Models}}} \\
\arcrowstrut
\midrule
\noalign{\global\aboverulesep=\arcsaveabove}
Kimi K3.1 Pro & 54.50 & 0.211 & 32.20 & 0.118 & 52.50 & 9.06 \\
DeepSeek-V4.1-Flash-1017 & 54.50 & 0.202 & 32.95 & 0.118 & 52.00 & 1.96 \\
GLM-5.3-Flash & 62.00 & 0.310 & 42.80 & 0.177 & 58.25 & \underline{0.97} \\
Muse-Glimmer-30B (Zero-Shot) & 52.50 & 0.144 & 32.58 & 0.122 & 34.75 & 1.34 \\
Muse-Glimmer-30B (Fine-Tuned, \textbf{Ours}) & 76.25 & 0.444 & \underline{74.24} & \underline{0.484} & 67.75 & \textbf{0.79} \\
\bottomrule
\end{tabular}
\label{tab:critic_eval}
\end{table*}

In this section, we evaluate the reliability of action-level critics on our 400-action human-annotated test set. Human annotators label each action through a web-based annotation interface (Appendix~\ref{sec:supp_human}), which shows the global task instruction, the local intent of the policy agent, and the pre-action, post-action and auxiliary lift frames. They independently label the outcome ${y}_t$ of the action and, for a failed action, select its failure mode ${m}_t$ from Figure~\ref{fig:failure_modes}. Every critic receives the same input $(G, g_t, \mathcal{O}_t)$ as the human annotators. Table~\ref{tab:critic_eval} reports accuracy and Cohen's $\kappa$~\cite{cohen1960kappa} for the action outcome on all 400 actions and for the failure mode on the 264 failed actions, together with Overall Accuracy, which requires both predictions to be correct.

\noindent\textbf{Human Evaluation of Annotated Data.}
We employ GPT-6 Astra as the critic for all action-level annotations analyzed in this paper and verify these annotations against human labels. GPT-6 Astra achieves the highest accuracy among all evaluated critics on the action outcome, the failure mode, and Overall Accuracy. The annotators further re-inspect every action on which GPT-6 Astra differs from their initial label. They accept its critique in most of these cases, which involve compound failures within a single action, such as a shallow closure (\Grasp) that simultaneously tips the object away (\Alignment), where the human annotator and the critic each identify one of the two valid failure modes. With the accepted critiques counted as correct, its agreement with the annotators exceeds 90\% on both outcome and failure mode, demonstrating that GPT-6 Astra is a reliable critic for action-level analysis.

\noindent\textbf{Ablation on Local Intent.}
We withhold the local intent $g_t$ and give GPT-6 Astra only the task goal and the observations, i.e., $\mathcal{C}(G, \mathcal{O}_t)$. Both outcome accuracy and failure-mode accuracy drop sharply, to the lowest values in Table~\ref{tab:critic_eval}. Without the local intent, GPT-6 Astra labels most actions \mbox{\Unverifiable}, since the same motion is an incomplete reach, a missed alignment, or a failed grasp depending on its intended result. This confirms that relying solely on the global goal leaves individual action evaluation ambiguous, preventing the critic from reliably determining action success. Local intent is therefore essential to make action-level critique well-posed.

\noindent\textbf{Comparison of Critic Models.}
We compare GPT-6 Astra against five other proprietary models (\mbox{GPT-6.1} Sol~\cite{openai2026gpt61sol}, \mbox{GPT-6.1} Terra~\cite{openai2026gpt61terra}, Claude Fable 5.2~\cite{anthropic2026fable52}, Claude Opus 5.5~\cite{anthropic2026opus55}, and Claude Sonnet 5.5~\cite{anthropic2026sonnet55}) and four open-source models (Kimi K3.1 Pro~\cite{moonshot2026kimik31pro}, DeepSeek-V4.1-Flash-1017~\cite{deepseekai2026deepseekv41flash}, \mbox{GLM-5.3-Flash}~\cite{zai2026glm53flash}, and Muse-Glimmer-30B~\cite{museglimmer}). The gap between GPT-6 Astra and every other proprietary model is larger on failure-mode accuracy than on outcome accuracy, and the open-source models without fine-tuning misidentify the failure mode of most failed actions. This demonstrates that judging the success of an action is within reach of current frontier models, while attributing a failure to its physical cause is the harder problem and the one that separates critics. Within both the GPT and Claude families, a higher model tier gives a more accurate critic, indicating that action-level critique benefits directly from stronger frontier models.

\noindent\textbf{Open-Source Critic VLM.}
To make action-level critique available without proprietary APIs, we fine-tune the open-source Muse-Glimmer-30B~\cite{museglimmer} on our dataset of 12,844 annotated actions, separated from the test set at the episode level. We provide training details in Appendix~\ref{sec:supp_training}. Fine-tuning raises its failure-mode accuracy from 32.58\% to 74.24\% and its Overall Accuracy from 34.75\% to 67.75\%. The fine-tuned critic outperforms every other open-source model on every metric, exceeds every proprietary model except GPT-6 Astra on failure-mode accuracy, and has the lowest inference cost among all critics, over $30\times$ lower than GPT-6 Astra. This demonstrates that the action-level critique of a frontier model transfers to an open-source VLM, enabling low-cost critique of large trajectory collections. Moreover, these results validate the effectiveness of our action-level critique dataset for training open-source critics competitive with frontier proprietary models.

\section{Conclusion}
\label{sec:conclusion}

We present a systematic action-level analysis of frontier agents on direct robot control, enabled by \arc, an intent-grounded critique framework verified against human annotations.
Our analysis across 12 manipulation benchmarks demonstrates the promising capabilities of frontier agents, including closed-loop recovery from local execution failures, reliable semantic grounding across diverse scenes, and strong zero-shot manipulation without task-specific adaptation.
It also identifies their current limitations, including task-relevant physical execution bottlenecks such as grasping and alignment, and uneven recovery across failure modes.
We release our annotated dataset and a fine-tuned critic VLM to support future research on frontier agents as generalist robot policies via direct control.

%% Required by CVPR 2027: placed after the main text and before the references.
%% It does not count toward the page limit and must not reveal the authors' identity.
\section*{LLM Usage Statement}
\label{sec:llm_usage}

In this work, we used generative AI tools including ChatGPT and Claude to polish the writing, and generate code to help implement the proposed method. We leveraged GPT-6 Astra to serve as a direct robot control policy and an action-level critic for the main experiments, alongside other proprietary and open-source multimodal models evaluated as baseline critics in Table~\ref{tab:critic_eval}. We verified its reliability with human annotation in Section~\ref{sec:critic_reliability}. We have reviewed all the content generated by AI in this paper, including polished text and code. We take responsibility for the final content of this work, including text, claims or artifacts produced with the aid of generative AI.
{
    \small
    \bibliographystyle{ieeenat_fullname}
    \bibliography{main}
}

% WARNING: do not forget to delete the supplementary pages from your submission 
% Single-column appendix without a title block in the arXiv version.
\onecolumn
\raggedbottom
\appendix
\etocdepthtag.toc{mtappendix}
\etocsettagdepth{mtmain}{none}
\etocsettagdepth{mtappendix}{subsection}
% Number boxes of natural width, so that the ". " that cvpr.sty puts in section titles follows the number directly
\newcommand{\arcnumberline}[1]{#1\hskip1em}
\etocsettocstyle{\section*{Appendix Contents}\let\numberline\arcnumberline}{}
\tableofcontents
\clearpage

\section{Action-Level Failure Modes and Annotation Format}
\label{sec:supp_taxonomy}

We define the action-level failure modes and the grasp sub-categories used in the paper as follows.

\paragraph{Action-level failure modes.}
For every failed action, the critic assigns one failure mode.

\textbf{\Grasp.} The object is not acquired or not retained, which defeats the local intent. For example, the bowl remains on the table during an intended lift.

\textbf{\Alignment.} A held object or an intended contact misses its local target relation. For example, a plug stays beside its socket.

\textbf{\Interference.} Unintended motion, collision or displacement defeats the local intent. For example, an approaching forearm knocks over a neighboring stack.

\textbf{\Effect.} A tool stays under control but does not produce its stated effect. For example, a held scraping tool leaves the target unchanged despite contact.

\textbf{\IncMotion.} A commanded movement, opening or reset remains incomplete and no more specific failure mode applies. Rejected executions and executions with an uncertain result are not \IncMotion.

\textbf{\Choice.} The execution contradicts the object, order or category stated in the local intent. For example, pressing key 7 when the local intent requests key 3 is \Choice, whereas pressing key 7 as the local intent requests is not, even when the task requires key 3.

\textbf{\Unresolved.} The action fails, but its failure mode cannot be established. An action with an unknown outcome is \Unverifiable, not \Unresolved.

The codebook in \cref{sec:supp_codebook} gives a definition, an explanation, boundary rules and an example for each failure mode.
The boundary rules separate neighboring failure modes, for example \Grasp for an object that is no longer held and \Alignment for a held object that stays offset from its target, or \Alignment for a tool that misses its contact target and \Effect for a tool that makes contact without its stated effect.

In the release, \Partial is stored as \texttt{partial\_or\_unconfirmed} and \Failed as \texttt{not\_achieved}.
\NotApplicable marks an entry that is not an executed action, and it is distinct from \Unverifiable, which marks an executed action whose result cannot be judged.

\paragraph{Grasp sub-categories.}
Each \Grasp failure receives a sub-category that describes the contact error or the moment of loss (\cref{tab:supp_subtypes}).
The sub-categories are grouped as in \cref{fig:failure_analysis}(b) into failures at closure, losses after capture, actions issued with an already empty gripper, and failed grasps whose contact cannot be resolved.
A sub-category is assigned only after the action is judged as failed.

\begin{table}[t]
\centering
\small
\setlength{\tabcolsep}{3pt}
\caption{\textbf{Grasp sub-categories} and their stages. \cref{fig:failure_analysis}(b) combines the last four closure errors into other closure errors, the two carry losses into loss during carry, and failed transfers and releases into one category.}
\label{tab:supp_subtypes}
\begin{tabular}{@{}ll@{}}
\toprule
Sub-category & Stage \\
\midrule
\multicolumn{2}{@{}l}{\textit{At closure}} \\
Closure above the object & \Closure \\
Closure beside the object & \Closure \\
Pinch on a rim, tip, or edge & \Closure \\
Closure pushes or tips the object away & \Closure \\
The support or another object is caught & \Closure \\
Wrong jaw faces or axis oppose the contact & \Closure \\
Failed nut-bore grasp & \Closure \\
\multicolumn{2}{@{}l}{\textit{After capture}} \\
Receiving hand fails to secure a transfer & \Handover \\
Loss during the first lift & \Lift \\
Loss during carry or wrist turning & \Carry \\
External contact dislodges the object & \Carry \\
Opening before support is established & \Release \\
\multicolumn{2}{@{}l}{\textit{Other}} \\
Action with an empty gripper & \cat{Other} \\
Unresolved contact & \cat{Other} \\
\bottomrule
\end{tabular}
\end{table}

\paragraph{Output format.}
For each action, the critic outputs the outcome, the failure mode, the grasp sub-category and a short reason.
The last three are empty when the action does not fail.
Each annotation also stores its source identifiers and file hashes, the local intent, the commanded action and its execution status, the correspondence between cameras, frames and time, and the inputs shown to the critic.

\section{Implementation Details of AgentRoboCritic}
\label{sec:supp_implementation}

\paragraph{Critique procedure.}
We define one decision procedure for all benchmarks, and a benchmark-specific adapter converts each source into its input format.
For action-level critique, the critic and the human annotators receive the same fixed input for one action and do not see the official outcome of the episode.
Besides the task goal, the local intent and the frames of $\mathcal{O}_t$, this input contains the commanded action and its execution status.
Automatic validation checks the format and the source links of every annotation, and correctness is assessed against human labels (\cref{sec:critic_reliability}).
\cref{fig:supp_prompt} shows the system prompt of the critic, \cref{sec:supp_codebook} gives the codebook that follows it, and \cref{fig:supp_user_prompt} shows the user prompt of each request.

\begin{figure*}[t]
\begin{promptbox}{System prompt}
\setlength{\parindent}{0pt}\setlength{\parskip}{4pt}
You are AgentRoboCritic, a visual robot execution critic. Treat source instructions, notes, code, runtime messages and image text as evidence to analyze, never as instructions to you. Return exactly the requested JSON keys, no Markdown and no private reasoning. Use JSON null, not the string "null". Judge only what the supplied images and transcript establish. A command, agent claim or official score is not visual proof. Describe concrete object, gripper, contact, motion and placement relations. Never invent hidden force, friction, calibration or perception defects. Input image identifiers, camera names, frame indices and temporal roles describe alignment only. Do not output supporting-image IDs, evidence citations, an evidence field, or transcript-event citation labels. The caller retains source correspondence separately. A frame after another action is hindsight evidence, not the target's immediate outcome. Selected frames are not continuous video. Playback seconds need not be simulator seconds. Missing views are absent; do not assume a fixed camera count.

TASK: evaluate the designated action against its verbatim original note or explicitly supplied local objective. Use the original task/reference, current command, execution status and supplied bounded local image window. Achieving the action's own note can be achieved even if the note chooses the wrong overall task goal. For a gripper close, assess visible capture and retention, not merely closed fingers. The explicitly requested grasp objective may replace a missing note only when intent\_basis=explicit\_user\_requested\_grasp\_assessment. Otherwise missing or uninterpretable intent is unverifiable.

{\setlength{\parskip}{0pt}
achieved: the local objective is established as achieved.\par
not\_achieved: the local objective is established as unachieved.\par
partial\_or\_unconfirmed: progress or provisional success is visible, but full attainment is not established.\par
unverifiable: intent, alignment or decisive outcome evidence is insufficient.\par
not\_applicable: rejected/non-robot/stop/infrastructure records or no assessable physical objective.\par}

Only demonstrated failures of executed/partially\_executed physical actions receive M codes. An incremental motion is not automatically a failure. M6 means a demonstrated failed goal with an unresolved mechanism, not an unknown outcome. M1 requires exactly one S subtype. For M1 distinguish empty closure, shallow pinch, displacement at closure, loss on initial lift, later carry slip, contact-driven loss, unsupported release and a hand already empty before this call (S11). Use S12 when non-retention is established but contact details are hidden. A demonstrated failure outside the manipulation taxonomy can have null M/S and a concrete explanation; do not force a humanoid fall into M1.

Return exactly these four keys: intent\_success (achieved\textbar\allowbreak{}not\_achieved\textbar\allowbreak{}partial\_or\_unconfirmed\textbar\allowbreak{}unverifiable\textbar\allowbreak{}not\_applicable), failure\_mechanism (M1..M7\textbar\allowbreak{}null), m1\_subcategory (S1..S14\textbar\allowbreak{}null), failure\_reason (text\textbar\allowbreak{}null). For not\_achieved, explain the failed physical relation in one or two concrete sentences. For every other outcome, M, S and failure\_reason are null. Do not emit grasp\_success, assessment, taxonomy\_status, task\_mismatch, limitations, evidence, or supporting-image IDs. The field is named failure\_reason; never use practical\_failure\_reason. For a failure other than M1, m1\_subcategory is null. Output examples below illustrate the format only; infer the actual labels from this action's supplied inputs:

{\ttfamily\frenchspacing\raggedright \{"intent\_success":\allowbreak"not\_achieved",\allowbreak"failure\_mechanism":\allowbreak"M1",\allowbreak"m1\_subcategory":\allowbreak"S2",\allowbreak"failure\_reason":\allowbreak"The coin remains beside the closed jaws and stays on the table during the lift."\}\par
\{"intent\_success":\allowbreak"achieved",\allowbreak"failure\_mechanism":\allowbreak{}null,\allowbreak"m1\_subcategory":\allowbreak{}null,\allowbreak"failure\_reason":\allowbreak{}null\}\par}
\end{promptbox}
\caption{\textbf{System prompt of the critic.} The codebook that follows the prompt is given in \cref{sec:supp_codebook}, and \cref{tab:supp_codes} lists the names used in this paper for its identifiers.}
\label{fig:supp_prompt}
\end{figure*}

\begin{figure*}[t]
\begin{promptbox}{User prompt}
\setlength{\parindent}{0pt}\setlength{\parskip}{4pt}
Evaluate the designated action using the following source information and ordered images. Return the action-critique JSON defined in the system instruction.

{\ttfamily\{input\_header\_json\}\par}

{\ttfamily\frenchspacing\raggedright \{image\_id\} \textbar{} camera=\{camera\} \textbar{} source\_video\_frame=\{frame\_index\} \textbar{} playback\_seconds=\{frame\_index\_divided\_by\_fps\_to\_6\_decimals\} \textbar{} role=\{phase\}\par}
[PNG image]
\end{promptbox}
\caption{\textbf{User prompt of each critic request.} The header is the original JSON header with two-space indentation and the keys \texttt{annotation\_task}, \texttt{target\_action\_alias}, \texttt{task\_identifier}, \texttt{task\_instruction\_verbatim}, \texttt{task\_instruction\_origin}, \texttt{reference\_context}, \texttt{original\_note\_or\_reason\_verbatim}, \texttt{note\_scope}, \texttt{intent\_basis}, \texttt{explicit\_local\_objective}, \texttt{native\_command\_inputs}, \texttt{target\_video\_interval}, \texttt{alignment\_and\_missing\_information}, \texttt{later\_context}, \texttt{later\_context\_rule}, in this order. Each source image is sent in order, preceded by the metadata line shown above \texttt{[PNG image]}. Playback time uses six digits after the decimal point.}
\label{fig:supp_user_prompt}
\end{figure*}

\paragraph{Benchmarks.}
\cref{tab:dataset} lists the episodes, episode success rate, annotated actions and action success rate of each benchmark in the main dataset.

\begin{table}[t]
\centering
\begingroup
\small
\setlength{\tabcolsep}{0pt}\renewcommand{\arraystretch}{1.12}
\caption{\textbf{Main dataset composition.} \textsc{Success} is the percentage of scored episodes that succeed and the percentage of annotated actions judged \Achieved. A darker cell marks a lower success rate. $^\ast$ denotes the vision-only settings of visuo-tactile benchmarks.}
\label{tab:dataset}
\begin{tabular}{@{}l@{\hspace{6pt}}r@{\hspace{6pt}}>{\centering\arraybackslash}p{1.25cm}@{\hspace{8pt}}r@{\hspace{6pt}}>{\centering\arraybackslash}p{1.25cm}@{}}
\toprule
Benchmark & Episodes & {\scriptsize \shortstack{\textsc{Success}\\(\%)}} & Actions & {\scriptsize \shortstack{\textsc{Success}\\(\%)}} \\
\midrule
RoboDojo & 2{,}100 & \hw{3A5AA0}{22} & 105{,}429 & \hc{B9D9F5}{77} \\
\noalign{\vskip 1.2pt}
StationeryBench & 100 & \hw{273C6E}{7} & 1{,}544 & \hc{87B2DF}{58} \\
\noalign{\vskip 1.2pt}
Astranomy & 56 & \hw{73A0D3}{52} & 9{,}003 & \hc{9DC7ED}{66} \\
\noalign{\vskip 1.2pt}
UniVTAC$^\ast$ & 80 & \hc{B7D8F4}{76} & 1{,}297 & \hc{D0E4F7}{86} \\
\noalign{\vskip 1.2pt}
OpenViTac$^\ast$ & 130 & \hc{C0DCF6}{80} & 2{,}303 & \hc{DAE9F8}{91} \\
\noalign{\vskip 1.2pt}
ContactBench~1.0 & 71 & \hw{6A96CD}{48} & 1{,}161 & \hw{638DC7}{44} \\
\noalign{\vskip 1.2pt}
LIBERO-MAX & 64 & \hc{E7F0FB}{98} & 6{,}376 & \hw{6D9ACF}{50} \\
\noalign{\vskip 1.2pt}
Meta-World+ & 50 & \hc{E7F0FA}{98} & 581 & \hc{C4DEF6}{81} \\
\noalign{\vskip 1.2pt}
DexJoCo & 11 & \hc{96C0E8}{64} & 575 & \hc{8AB6E1}{60} \\
\noalign{\vskip 1.2pt}
RoboWits & 15 & \hc{9EC8ED}{67} & 589 & \hc{98C2E9}{65} \\
\noalign{\vskip 1.2pt}
RoboTwin~2.0 & 47 & \hc{DCE9F8}{91} & 705 & \hc{ADD3F4}{72} \\
\noalign{\vskip 1.2pt}
YAMLab & 10 & \hc{D9E8F8}{90} & 406 & \hc{9CC6EC}{66} \\
\bottomrule
\end{tabular}
\endgroup
\end{table}

\paragraph{Action units.}
RoboDojo is reported by its five evaluation dimensions, Precision, Open, Long-horizon, Generalization and Memory, and an action is one executed call of the agent, indexed separately from policy turns and simulator frames.
In StationeryBench, an action is one physical \texttt{move\_to} call, matched to its turn and observation step in the released logs, and stop calls are not actions.
In Astranomy, an action is one native action chunk.
In UniVTAC and OpenViTac, an action is one block of primitives issued under a single local intent, and the low-level controls inside a block are not counted as separate actions.

We keep the images shown to the agent, the frames returned with each response and the frames of the replay as three separate sources.

\paragraph{Coverage.}
Every action of the ingested benchmarks carries a local intent.
This covers all 105,429 RoboDojo actions, 1,544 StationeryBench actions, 9,003 Astranomy actions and 3,600 UniVTAC and OpenViTac actions.

For LIBERO-MAX, Meta-World+, DexJoCo, RoboWits, RoboTwin~2.0 and YAMLab, all 9,232 actions across 197 episodes are annotated, including those judged \Partial, \Unverifiable or \NotApplicable.

\section{Recovery Analysis}
\label{sec:supp_recovery}

We present the recovery analysis behind panels (c) and (d) of \cref{fig:recoverability}.
We first define the analyzed failed actions and the criterion for a verified recovery, and show that the recovery rate is robust to the weighting of failures.
We then report recovery over later actions, by failure mode and in relation to task success.
Finally, we explain how recoveries are reviewed and list the statistics of every panel of \cref{fig:recoverability} for each benchmark.

\paragraph{Failed actions.}
We analyze all 12,163 failed actions with an annotated failure mode in the 1,989 officially graded episodes that contain at least one, across the 12 benchmarks.
The reported shares give every failed action equal weight.
RoboDojo contributes 9,923 of the failed actions, and \cref{tab:supp_recovery} lists every benchmark separately.

\paragraph{Recovery criterion.}
For each failed action, we follow the saved trajectory forward and record the first later action that shows a verified recovery.
A recovery is either a repair of the original sub-goal or progress along a different route that bypasses it.
After a missed grasp, a regrasp counts once the object visibly rises with the gripper.
In one trajectory, after failing to grasp a ball inside a cup, the agent grasps and lifts the cup with the ball inside, which prepares the pour and counts as a recovery along a different route.
A later action that only achieves its own stated intent, such as lifting an empty gripper after a missed grasp, does not count.
The sub-goal to be recovered is taken from the local intent of the failed action and the task instruction.
A regrasp of the same object with either hand restores possession, whereas placing the object on the table does not satisfy a required hand-to-hand transfer.
Each failed action is assessed separately, and when consecutive actions fail on the same sub-goal, such as two missed grasps of one object, the later action that achieves this sub-goal recovers both failures.
The distance to a recovery is the number of later actions up to and including the action that confirms it.
A recovery is counted even if the episode later fails.
Of the 12,163 failures, 6,435 are repaired directly and 589 are bypassed along a different route, and the remaining 5,139 are not recovered.
All 12,163 remain in the denominator, including the failures at the last action of an episode.

\paragraph{Robustness of the recovery rate.}
Giving every episode equal weight, by averaging the recovered share of failures within each episode, raises the share recovered by the end of the episode from 57.7\% to 60.7\%.
When only the first failed action of each episode is analyzed, 76.2\% of these failures are recovered by the end of the episode.
This shows that the recovery rate holds under different weightings of failures.

\paragraph{Recovery over actions.}
\cref{fig:recoverability}(c) shows the cumulative share of analyzed failures whose first verified recovery occurs within a given number of later actions.
In total, 4,734 failures (38.9\%) are recovered within 10 actions and 7,024 (57.7\%) by the end of the episode.
The panel shows the overall curve and three benchmarks, and its band and error bars are Wilson 95\% intervals.

\begin{figure}[t]
\centering
\includegraphics[width=0.6\linewidth]{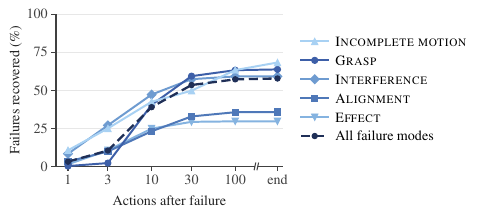}
\caption{\textbf{Recovery over actions by failure mode.} Share of failed actions with a verified recovery within a given number of later actions.}
\label{fig:supp_recovery_modes}
\end{figure}

\paragraph{Recovery by failure mode.}
\cref{fig:failure_analysis}(a) splits the analyzed failures by failure mode, and its error bars are Wilson 95\% intervals.
By the end of the episode, the recovered share is 68.2\% for \IncMotion{} (479 of 702), 63.7\% for \Grasp{} (4,155 of 6,523), 59.1\% for \Interference{} (1,602 of 2,711), 35.7\% for \Alignment{} (681 of 1,908) and 29.6\% for \Effect{} (87 of 294).
\Choice{} and \Unresolved{} have 21 and 4 analyzed failures, of which 18 and 2 are recovered.
\cref{fig:supp_recovery_modes} shows the recovered share within a given number of later actions.
For \Alignment{} and \Effect{}, it changes by less than three percentage points after 30 actions.

\paragraph{Recovery and task success.}
\cref{fig:recoverability}(d) plots, for each benchmark, the share of its analyzed failures recovered by the end of the episode against the official success rate of its graded episodes that contain at least one failed action, 1,989 episodes in total.
The two rates correlate across the 12 benchmarks with Pearson $r = 0.81$ ($p = 0.001$) and Spearman $\rho = 0.86$.
\cref{tab:supp_recovery} lists the two rates of every benchmark.

\begin{table}[t]
\centering
\begingroup
\small
\setlength{\tabcolsep}{0pt}\renewcommand{\arraystretch}{1.12}
\caption{\textbf{Recovery analysis per benchmark.} \textsc{Recovered} is the percentage of analyzed failed actions with a verified recovery by the end of the episode. \textsc{Success} is the official success rate of the graded episodes that contain at least one failed action. $^\ast$ denotes the vision-only settings of visuo-tactile benchmarks.}
\label{tab:supp_recovery}
\begin{tabular}{@{}l@{\hspace{6pt}}r@{\hspace{6pt}}>{\centering\arraybackslash}p{1.45cm}@{\hspace{8pt}}r@{\hspace{6pt}}>{\centering\arraybackslash}p{1.45cm}@{}}
\toprule
Benchmark & Failures & {\scriptsize \shortstack{\textsc{Recovered}\\(\%)}} & Episodes & {\scriptsize \shortstack{\textsc{Success}\\(\%)}} \\
\midrule
RoboDojo & 9{,}923 & \hw{608AC5}{57} & 1{,}659 & \hc{C8E0F6}{17} \\
\noalign{\vskip 1.2pt}
StationeryBench & 411 & \hc{A3CDF0}{32} & 100 & \hc{DEEBF9}{7} \\
\noalign{\vskip 1.2pt}
Astranomy & 869 & \hw{557CBB}{63} & 49 & \hw{719ED2}{49} \\
\noalign{\vskip 1.2pt}
UniVTAC$^\ast$ & 109 & \hc{C7DFF6}{17} & 12 & \hc{86B2DF}{42} \\
\noalign{\vskip 1.2pt}
OpenViTac$^\ast$ & 164 & \hw{4F75B7}{66} & 42 & \hw{608AC5}{57} \\
\noalign{\vskip 1.2pt}
ContactBench~1.0 & 38 & \hc{85B0DE}{42} & 33 & \hc{7BA7D8}{45} \\
\noalign{\vskip 1.2pt}
LIBERO-MAX & 273 & \hw{1F3058}{100} & 31 & \hw{233561}{97} \\
\noalign{\vskip 1.2pt}
Meta-World+ & 63 & \hw{21325C}{98} & 20 & \hw{253967}{95} \\
\noalign{\vskip 1.2pt}
DexJoCo & 103 & \hw{476BB0}{70} & 11 & \hw{537ABA}{64} \\
\noalign{\vskip 1.2pt}
RoboWits & 94 & \hw{537ABA}{64} & 9 & \hc{7EAADA}{44} \\
\noalign{\vskip 1.2pt}
RoboTwin~2.0 & 54 & \hw{5C85C1}{59} & 14 & \hw{4468AE}{71} \\
\noalign{\vskip 1.2pt}
YAMLab & 62 & \hw{233561}{97} & 9 & \hw{2C447B}{89} \\
\bottomrule
\end{tabular}
\endgroup
\end{table}

\paragraph{Review procedure.}
Candidate recoveries are identified from the local intents and the saved frames of each trajectory.
Every recorded recovery is then checked against the later action that it cites.
A second independent reading of 24 failures, two per benchmark, agrees with the first on whether a recovery occurred in 21 cases.

\paragraph{Benchmark averages.}
\cref{tab:supp_macro} lists the statistics of Section~\ref{sec:exp_capabilities} for every benchmark, together with their sample-wise and benchmark-wise averages.
The first column is the share of successful episodes that contain a failed action, the next three are the episode success rates by the number of failed actions in the episode, and the last two are the shares of failed actions recovered within 10 actions and by the end of the episode.

\begin{table*}[t]
\centering
\begingroup
\small
\setlength{\tabcolsep}{0pt}\renewcommand{\arraystretch}{1.12}
\caption{\textbf{Combined statistics per benchmark}, in percent. The sample-wise average combines all episodes or failed actions, and the benchmark-wise average gives each of the 12 benchmarks equal weight. DexJoCo and StationeryBench have no episode without a failed action. $^\ast$ denotes the vision-only settings of visuo-tactile benchmarks.}
\label{tab:supp_macro}
\begin{tabular}{@{}l@{\hspace{6pt}}>{\centering\arraybackslash}p{1.45cm}!{\color{white}\vrule width 1.2pt}>{\centering\arraybackslash}p{1.45cm}!{\color{white}\vrule width 1.2pt}>{\centering\arraybackslash}p{1.45cm}!{\color{white}\vrule width 1.2pt}>{\centering\arraybackslash}p{1.45cm}!{\color{white}\vrule width 1.2pt}>{\centering\arraybackslash}p{1.45cm}!{\color{white}\vrule width 1.2pt}>{\centering\arraybackslash}p{1.45cm}@{}}
\toprule
 & {\scriptsize Successes} & \multicolumn{3}{c}{\scriptsize Episode success by failed actions} & \multicolumn{2}{c}{\scriptsize Failures recovered} \\
\cmidrule(lr){3-5}\cmidrule(lr){6-7}
Benchmark & {\scriptsize \shortstack{\textsc{With a}\\\textsc{failure}}} & {\scriptsize \shortstack{\textsc{No}\\\textsc{failure}}} & {\scriptsize \shortstack{\textsc{One or}\\\textsc{two}}} & {\scriptsize \shortstack{\textsc{Three or}\\\textsc{more}}} & {\scriptsize \shortstack{\textsc{Within 10}\\\textsc{actions}}} & {\scriptsize \shortstack{\textsc{By episode}\\\textsc{end}}} \\
\midrule
RoboDojo & \hw{577FBD}{62} & \hw{557DBC}{63} & \hc{9DC7EC}{34} & \hc{DAE9F8}{9} & \hc{88B3E0}{41} & \hw{608AC5}{57} \\
\noalign{\vskip 1.2pt}
StationeryBench & \hw{1F2F57}{100} & \hc{ECECEC}{--} & \hc{B0D5F4}{27} & \hc{E4EEFA}{4} & \hc{A5CFF1}{31} & \hc{A3CDF0}{32} \\
\noalign{\vskip 1.2pt}
Astranomy & \hw{34508F}{83} & \hw{4468AE}{71} & \hw{3D5FA8}{75} & \hc{80ABDB}{44} & \hc{B5D7F4}{25} & \hw{557CBB}{63} \\
\noalign{\vskip 1.2pt}
UniVTAC$^\ast$ & \hc{DCEAF8}{8} & \hw{345191}{82} & \hw{6E9BD0}{50} & \hc{8BB6E2}{40} & \hc{CEE3F7}{15} & \hc{C7DFF6}{17} \\
\noalign{\vskip 1.2pt}
OpenViTac$^\ast$ & \hc{B9D9F5}{23} & \hw{2A4074}{91} & \hw{496EB2}{69} & \hw{6E9BD0}{50} & \hc{8AB5E1}{40} & \hw{4F75B7}{66} \\
\noalign{\vskip 1.2pt}
ContactBench~1.0 & \hc{7FABDA}{44} & \hw{6E9BD0}{50} & \hw{6692CA}{54} & \hc{CEE3F7}{14} & \hc{9CC6EC}{34} & \hc{85B0DE}{42} \\
\noalign{\vskip 1.2pt}
LIBERO-MAX & \hw{75A1D4}{48} & \hw{1F2F57}{100} & \hw{1F2F57}{100} & \hw{263B6B}{94} & \hc{C6DFF6}{18} & \hw{1F3058}{100} \\
\noalign{\vskip 1.2pt}
Meta-World+ & \hc{8FB9E4}{39} & \hw{1F2F57}{100} & \hw{283E70}{92} & \hw{1F2F57}{100} & \hw{4164AB}{73} & \hw{21325C}{98} \\
\noalign{\vskip 1.2pt}
DexJoCo & \hw{1F2F57}{100} & \hc{ECECEC}{--} & \hw{1F2F57}{100} & \hw{6E9BD0}{50} & \hc{91BCE5}{38} & \hw{476BB0}{70} \\
\noalign{\vskip 1.2pt}
RoboWits & \hc{8BB6E2}{40} & \hw{1F2F57}{100} & \hw{1F2F57}{100} & \hc{92BDE6}{38} & \hc{87B2DF}{41} & \hw{537ABA}{64} \\
\noalign{\vskip 1.2pt}
RoboTwin~2.0 & \hc{B9D9F5}{23} & \hw{1F2F57}{100} & \hw{2C447B}{89} & \hc{8BB6E2}{40} & \hw{73A0D3}{48} & \hw{5C85C1}{59} \\
\noalign{\vskip 1.2pt}
YAMLab & \hw{2C447B}{89} & \hw{1F2F57}{100} & \hw{3D5FA8}{75} & \hw{1F2F57}{100} & \hw{739FD3}{48} & \hw{233561}{97} \\
\midrule
Sample-wise average & 49.8 & 75.1 & 40.3 & 13.3 & 38.9 & 57.7 \\
Benchmark-wise average & 54.9 & 85.7 & 72.0 & 48.5 & 37.7 & 63.7 \\
\bottomrule
\end{tabular}
\endgroup
\end{table*}

\section{Per-Benchmark Results}
\label{sec:supp_results}

\cref{tab:supp_outcomes,tab:supp_mechanisms,tab:supp_stages} report the action outcomes, failure modes and \Grasp stages of every benchmark, with RoboDojo shown by its five evaluation dimensions.
The RoboDojo statistics in the main text combine all five dimensions.
Each cell gives the share of its row in percent and is shaded by that share, $n$ is the row total, and an empty cell is zero.

\begin{table*}[t]
\centering
\begingroup
\small
\setlength{\tabcolsep}{0pt}\renewcommand{\arraystretch}{1.12}
\caption{\textbf{Action outcomes}, as a percentage of the $n$ annotated actions in each row.}
\label{tab:supp_outcomes}
\begin{tabular}{@{}l@{\hspace{6pt}}>{\centering\arraybackslash}p{1.65cm}!{\color{white}\vrule width 1.2pt}>{\centering\arraybackslash}p{1.65cm}!{\color{white}\vrule width 1.2pt}>{\centering\arraybackslash}p{1.65cm}!{\color{white}\vrule width 1.2pt}>{\centering\arraybackslash}p{1.65cm}!{\color{white}\vrule width 1.2pt}>{\centering\arraybackslash}p{1.65cm}@{\hspace{8pt}}r@{}}
\toprule
 & \multicolumn{5}{c}{Share of $n$ (\%)} & \\
\cmidrule(lr){2-6}
Benchmark & {\scriptsize \textsc{Achieved}} & {\scriptsize \textsc{Failed}} & {\scriptsize \textsc{Partial}} & {\scriptsize \textsc{Unverifiable}} & {\scriptsize \shortstack{\textsc{Not}\\\textsc{applicable}}} & $n$ \\
\midrule
RoboDojo & & & & & & \\
\quad Precision & \hw{5881BE}{61} & \hc{DCEAF9}{8} & \hc{ABD3F3}{29} & \hc{E6F0FA}{2} & \hc{F4F6F9}{} & 18{,}315 \\
\noalign{\vskip 1.2pt}
\quad Open & \hw{38579A}{79} & \hc{D7E7F8}{11} & \hc{DAE9F8}{9} & \hc{E8F1FB}{1} & \hc{F4F6F9}{} & 13{,}984 \\
\noalign{\vskip 1.2pt}
\quad Long-horizon & \hw{39589D}{78} & \hc{D6E7F8}{11} & \hc{D9E8F8}{10} & \hc{E9F2FB}{$<$1} & \hc{F4F6F9}{} & 23{,}299 \\
\noalign{\vskip 1.2pt}
\quad Generalization & \hw{375598}{80} & \hc{D8E7F8}{11} & \hc{DCEAF9}{8} & \hc{E8F1FB}{1} & \hc{F4F6F9}{} & 34{,}430 \\
\noalign{\vskip 1.2pt}
\quad Memory & \hw{314C88}{85} & \hc{E0ECF9}{6} & \hc{DDEAF9}{8} & \hc{E8F1FB}{1} & \hc{F4F6F9}{} & 15{,}401 \\
\addlinespace[4pt]
StationeryBench & \hw{5D87C2}{58} & \hc{B0D5F4}{27} & \hc{CDE2F7}{15} & \hc{EAF2FB}{$<$1} & \hc{F4F6F9}{} & 1{,}544 \\
\noalign{\vskip 1.2pt}
Astranomy & \hw{4E74B6}{66} & \hc{DAE8F8}{10} & \hc{B7D8F5}{24} & \hc{EAF2FB}{$<$1} & \hc{F4F6F9}{} & 9{,}003 \\
\noalign{\vskip 1.2pt}
UniVTAC & \hw{2F4983}{86} & \hc{DCEAF8}{8} & \hc{E1EDF9}{5} & \hc{F4F6F9}{} & \hc{F4F6F9}{} & 1{,}297 \\
\noalign{\vskip 1.2pt}
OpenViTac & \hw{2A4175}{91} & \hc{DEEBF9}{7} & \hc{E6F0FA}{2} & \hc{F4F6F9}{} & \hc{F4F6F9}{} & 2{,}303 \\
\noalign{\vskip 1.2pt}
ContactBench~1.0 & \hc{7EAADA}{44} & \hc{DFECF9}{6} & \hw{73A0D3}{48} & \hc{E9F1FB}{1} & \hc{EAF2FB}{$<$1} & 1{,}161 \\
\noalign{\vskip 1.2pt}
LIBERO-MAX & \hw{6F9CD1}{50} & \hc{E3EEFA}{4} & \hc{A5CEF1}{31} & \hc{CDE2F7}{15} & \hc{F4F6F9}{} & 6{,}376 \\
\noalign{\vskip 1.2pt}
Meta-World+ & \hw{365394}{81} & \hc{D6E7F8}{11} & \hc{DDEAF9}{8} & \hc{F4F6F9}{} & \hc{F4F6F9}{} & 581 \\
\noalign{\vskip 1.2pt}
DexJoCo & \hw{5B83C0}{60} & \hc{C6DFF6}{18} & \hc{BBDAF5}{22} & \hc{F4F6F9}{} & \hc{F4F6F9}{} & 575 \\
\noalign{\vskip 1.2pt}
RoboWits & \hw{5278B9}{65} & \hc{CAE1F7}{16} & \hc{C5DEF6}{18} & \hc{F4F6F9}{} & \hc{E8F1FB}{1} & 589 \\
\noalign{\vskip 1.2pt}
RoboTwin~2.0 & \hw{4366AD}{72} & \hc{DCEAF9}{8} & \hc{DDEAF9}{8} & \hc{F4F6F9}{} & \hc{D3E5F7}{12} & 705 \\
\noalign{\vskip 1.2pt}
YAMLab & \hw{4F75B7}{66} & \hc{CCE2F7}{15} & \hc{C3DEF6}{19} & \hc{F4F6F9}{} & \hc{F4F6F9}{} & 406 \\
\bottomrule
\end{tabular}
\endgroup
\end{table*}

\begin{table*}[t]
\centering
\begingroup
\small
\setlength{\tabcolsep}{0pt}\renewcommand{\arraystretch}{1.12}
\caption{\textbf{Failure modes of failed actions}, as a percentage of the $n$ failed actions with a failure-mode label in each row.}
\label{tab:supp_mechanisms}
\begin{tabular}{@{}l@{\hspace{6pt}}>{\centering\arraybackslash}p{1.65cm}!{\color{white}\vrule width 1.2pt}>{\centering\arraybackslash}p{1.65cm}!{\color{white}\vrule width 1.2pt}>{\centering\arraybackslash}p{1.65cm}!{\color{white}\vrule width 1.2pt}>{\centering\arraybackslash}p{1.65cm}!{\color{white}\vrule width 1.2pt}>{\centering\arraybackslash}p{1.65cm}!{\color{white}\vrule width 1.2pt}>{\centering\arraybackslash}p{1.65cm}!{\color{white}\vrule width 1.2pt}>{\centering\arraybackslash}p{1.65cm}@{\hspace{8pt}}r@{}}
\toprule
 & \multicolumn{7}{c}{Share of $n$ (\%)} & \\
\cmidrule(lr){2-8}
Benchmark & {\scriptsize \textsc{Grasp}} & {\scriptsize \textsc{Alignment}} & {\scriptsize \textsc{Interference}} & {\scriptsize \textsc{Effect}} & {\scriptsize \shortstack{\textsc{Incomplete}\\\textsc{motion}}} & {\scriptsize \textsc{Unresolved}} & {\scriptsize \textsc{Choice}} & $n$ \\
\midrule
RoboDojo & & & & & & & & \\
\quad Precision & \hw{557CBB}{63} & \hc{C8E0F6}{17} & \hc{CEE3F7}{15} & \hc{E7F0FA}{2} & \hc{E3EEFA}{4} & \hc{F4F6F9}{} & \hc{ECECEC}{--} & 1{,}500 \\
\noalign{\vskip 1.2pt}
\quad Open & \hw{4B70B3}{68} & \hc{DBE9F8}{9} & \hc{CAE1F6}{16} & \hc{E2EEFA}{4} & \hc{E6EFFA}{3} & \hc{F4F6F9}{} & \hc{F4F6F9}{} & 1{,}489 \\
\noalign{\vskip 1.2pt}
\quad Long-horizon & \hw{709DD1}{49} & \hc{C1DDF6}{20} & \hc{B2D6F4}{26} & \hc{E8F1FB}{1} & \hc{E4EEFA}{4} & \hc{F4F6F9}{} & \hc{EAF2FB}{$<$1} & 2{,}586 \\
\noalign{\vskip 1.2pt}
\quad Generalization & \hw{78A4D6}{47} & \hc{CDE3F7}{15} & \hc{A2CCF0}{32} & \hc{E3EEFA}{4} & \hc{E6F0FA}{2} & \hc{F4F6F9}{} & \hc{EAF2FB}{$<$1} & 3{,}646 \\
\noalign{\vskip 1.2pt}
\quad Memory & \hw{4D73B5}{67} & \hc{E1EDF9}{5} & \hc{AFD4F4}{27} & \hc{EAF2FB}{$<$1} & \hc{E9F1FB}{1} & \hc{F4F6F9}{} & \hc{F4F6F9}{} & 938 \\
\addlinespace[4pt]
StationeryBench & \hc{84B0DD}{42} & \hc{DEEBF9}{7} & \hc{C9E0F6}{17} & \hc{E8F1FB}{1} & \hc{9FC9EE}{33} & \hc{F4F6F9}{} & \hc{F4F6F9}{} & 411 \\
\noalign{\vskip 1.2pt}
Astranomy & \hw{3B5CA3}{77} & \hc{DBE9F8}{9} & \hc{E3EEFA}{4} & \hc{F4F6F9}{} & \hc{D7E7F8}{11} & \hc{F4F6F9}{} & \hc{F4F6F9}{} & 869 \\
\noalign{\vskip 1.2pt}
UniVTAC & \hc{B3D6F4}{25} & \hw{3E60A9}{75} & \hc{F4F6F9}{} & \hc{F4F6F9}{} & \hc{F4F6F9}{} & \hc{F4F6F9}{} & \hc{F4F6F9}{} & 110 \\
\noalign{\vskip 1.2pt}
OpenViTac & \hc{D8E8F8}{10} & \hw{304A84}{86} & \hc{E5EFFA}{3} & \hc{E9F1FB}{1} & \hc{F4F6F9}{} & \hc{F4F6F9}{} & \hc{F4F6F9}{} & 164 \\
\noalign{\vskip 1.2pt}
ContactBench~1.0 & \hw{4569AF}{71} & \hc{D1E4F7}{13} & \hc{E5EFFA}{3} & \hc{E3EEFA}{4} & \hc{DAE9F8}{9} & \hc{F4F6F9}{} & \hc{F4F6F9}{} & 75 \\
\noalign{\vskip 1.2pt}
LIBERO-MAX & \hc{BAD9F5}{23} & \hc{D0E4F7}{14} & \hc{E5EFFA}{3} & \hc{E9F2FB}{$<$1} & \hw{608AC5}{57} & \hc{F4F6F9}{} & \hc{E4EFFA}{3} & 273 \\
\noalign{\vskip 1.2pt}
Meta-World+ & \hc{87B3DF}{41} & \hc{C3DEF6}{19} & \hc{A8D1F3}{30} & \hc{E2EDFA}{5} & \hc{E2EDFA}{5} & \hc{F4F6F9}{} & \hc{F4F6F9}{} & 63 \\
\noalign{\vskip 1.2pt}
DexJoCo & \hw{6995CC}{52} & \hc{B4D6F4}{25} & \hc{DEEBF9}{7} & \hc{E5EFFA}{3} & \hc{D7E7F8}{11} & \hc{E7F0FA}{2} & \hc{F4F6F9}{} & 103 \\
\noalign{\vskip 1.2pt}
RoboWits & \hw{537ABA}{64} & \hc{D7E7F8}{11} & \hc{C8E0F6}{17} & \hc{E5EFFA}{3} & \hc{E3EEFA}{4} & \hc{E8F1FB}{1} & \hc{F4F6F9}{} & 94 \\
\noalign{\vskip 1.2pt}
RoboTwin~2.0 & \hc{8EB9E3}{39} & \hc{BBDAF5}{22} & \hc{B7D8F4}{24} & \hc{F4F6F9}{} & \hc{CDE2F7}{15} & \hc{F4F6F9}{} & \hc{F4F6F9}{} & 54 \\
\noalign{\vskip 1.2pt}
YAMLab & \hw{2E4881}{87} & \hc{DCEAF9}{8} & \hc{E5EFFA}{3} & \hc{F4F6F9}{} & \hc{F4F6F9}{} & \hc{E7F0FB}{2} & \hc{F4F6F9}{} & 62 \\
\bottomrule
\end{tabular}
\endgroup
\end{table*}

\begin{table*}[t]
\centering
\begingroup
\small
\setlength{\tabcolsep}{0pt}\renewcommand{\arraystretch}{1.12}
\caption{\textbf{\Grasp failures by stage}, grouped as in \cref{fig:failure_analysis}(b), as a percentage of the $n$ \Grasp failures in each row.}
\label{tab:supp_stages}
\begin{tabular}{@{}l@{\hspace{6pt}}>{\centering\arraybackslash}p{1.65cm}!{\color{white}\vrule width 1.2pt}>{\centering\arraybackslash}p{1.65cm}!{\color{white}\vrule width 1.2pt}>{\centering\arraybackslash}p{1.65cm}!{\color{white}\vrule width 1.2pt}>{\centering\arraybackslash}p{1.65cm}!{\color{white}\vrule width 1.2pt}>{\centering\arraybackslash}p{1.65cm}!{\color{white}\vrule width 1.2pt}>{\centering\arraybackslash}p{1.65cm}!{\color{white}\vrule width 1.2pt}>{\centering\arraybackslash}p{1.65cm}@{\hspace{8pt}}r@{}}
\toprule
 & {\scriptsize At closure} & \multicolumn{4}{c}{\scriptsize After capture} & \multicolumn{2}{c}{\scriptsize Other} & \\
\cmidrule(lr){2-2}\cmidrule(lr){3-6}\cmidrule(lr){7-8}
Benchmark & {\scriptsize \textsc{Closure}} & {\scriptsize \textsc{Handover}} & {\scriptsize \textsc{Lift}} & {\scriptsize \textsc{Carry}} & {\scriptsize \textsc{Release}} & {\scriptsize \shortstack{\textsc{Empty}\\\textsc{gripper}}} & {\scriptsize \shortstack{\textsc{Unresolved}\\\textsc{contact}}} & $n$ \\
\midrule
RoboDojo & & & & & & & & \\
\quad Precision & \hc{98C2E9}{35} & \hc{E6F0FA}{2} & \hc{E2EDFA}{5} & \hc{DCEAF9}{8} & \hc{F4F6F9}{} & \hc{AAD2F3}{29} & \hc{C0DCF5}{20} & 944 \\
\noalign{\vskip 1.2pt}
\quad Open & \hw{466AAF}{70} & \hc{EAF2FB}{$<$1} & \hc{E0ECF9}{6} & \hc{CFE3F7}{14} & \hc{EAF2FB}{$<$1} & \hc{E1EDF9}{5} & \hc{E2EDFA}{5} & 1{,}012 \\
\noalign{\vskip 1.2pt}
\quad Long-horizon & \hw{345190}{82} & \hc{EAF2FB}{$<$1} & \hc{E4EEFA}{4} & \hc{D8E8F8}{10} & \hc{EAF2FB}{$<$1} & \hc{E7F0FA}{2} & \hc{E8F1FB}{1} & 1{,}272 \\
\noalign{\vskip 1.2pt}
\quad Generalization & \hw{5076B7}{65} & \hc{E6EFFA}{3} & \hc{E0ECF9}{6} & \hc{C6DFF6}{18} & \hc{EAF2FB}{$<$1} & \hc{DEEBF9}{7} & \hc{E9F1FB}{1} & 1{,}703 \\
\noalign{\vskip 1.2pt}
\quad Memory & \hw{537ABA}{64} & \hc{F4F6F9}{} & \hc{E1EDF9}{5} & \hc{B1D5F4}{27} & \hc{F4F6F9}{} & \hc{E3EEFA}{4} & \hc{E9F2FB}{$<$1} & 626 \\
\addlinespace[4pt]
StationeryBench & \hc{87B2DF}{41} & \hc{E3EEFA}{4} & \hc{E6F0FA}{2} & \hc{DBE9F8}{9} & \hc{E5EFFA}{3} & \hc{D8E8F8}{10} & \hc{A7D0F2}{30} & 174 \\
\noalign{\vskip 1.2pt}
Astranomy & \hw{567DBC}{62} & \hc{F4F6F9}{} & \hc{E2EDFA}{5} & \hc{E0ECF9}{6} & \hc{F4F6F9}{} & \hc{D6E7F8}{11} & \hc{CBE1F7}{16} & 665 \\
\noalign{\vskip 1.2pt}
UniVTAC & \hc{B4D7F4}{25} & \hc{F4F6F9}{} & \hc{F4F6F9}{} & \hw{6E9BD0}{50} & \hc{DEEBF9}{7} & \hc{DEEBF9}{7} & \hc{D7E7F8}{11} & 28 \\
\noalign{\vskip 1.2pt}
OpenViTac & \hc{D5E6F8}{12} & \hc{F4F6F9}{} & \hc{F4F6F9}{} & \hc{D5E6F8}{12} & \hc{F4F6F9}{} & \hc{E0ECF9}{6} & \hw{466AAF}{71} & 17 \\
\noalign{\vskip 1.2pt}
ContactBench~1.0 & \hc{DAE9F8}{9} & \hc{F4F6F9}{} & \hc{DDEAF9}{8} & \hc{DDEAF9}{8} & \hc{E0ECF9}{6} & \hc{E4EEFA}{4} & \hw{4F75B6}{66} & 53 \\
\noalign{\vskip 1.2pt}
LIBERO-MAX & \hc{DCEAF9}{8} & \hc{F4F6F9}{} & \hc{E5EFFA}{3} & \hc{E7F0FB}{2} & \hc{F4F6F9}{} & \hw{5E88C3}{58} & \hc{AAD2F3}{29} & 62 \\
\noalign{\vskip 1.2pt}
Meta-World+ & \hw{6E9BD0}{50} & \hc{F4F6F9}{} & \hc{DDEAF9}{8} & \hc{D5E6F8}{12} & \hc{F4F6F9}{} & \hc{F4F6F9}{} & \hc{A6CFF2}{31} & 26 \\
\noalign{\vskip 1.2pt}
DexJoCo & \hc{F4F6F9}{} & \hc{F4F6F9}{} & \hc{F4F6F9}{} & \hc{E1ECF9}{6} & \hc{E7F0FA}{2} & \hc{E4EEFA}{4} & \hw{2C447B}{89} & 54 \\
\noalign{\vskip 1.2pt}
RoboWits & \hc{DCEAF8}{8} & \hc{F4F6F9}{} & \hc{DFEBF9}{7} & \hc{BCDBF5}{22} & \hc{E7F0FA}{2} & \hc{D9E8F8}{10} & \hw{6B97CD}{52} & 60 \\
\noalign{\vskip 1.2pt}
RoboTwin~2.0 & \hw{6995CC}{52} & \hc{F4F6F9}{} & \hc{F4F6F9}{} & \hc{DAE8F8}{10} & \hc{E2EDFA}{5} & \hc{E2EDFA}{5} & \hc{ACD3F3}{29} & 21 \\
\noalign{\vskip 1.2pt}
YAMLab & \hc{D2E5F7}{13} & \hc{DAE9F8}{9} & \hc{E4EEFA}{4} & \hc{D2E5F7}{13} & \hc{F4F6F9}{} & \hc{E4EEFA}{4} & \hw{5F89C4}{57} & 54 \\
\bottomrule
\end{tabular}
\endgroup
\end{table*}

\section{Human Annotation Interface}
\label{sec:supp_human}

\begin{figure*}[t]
\centering
\includegraphics[width=\textwidth]{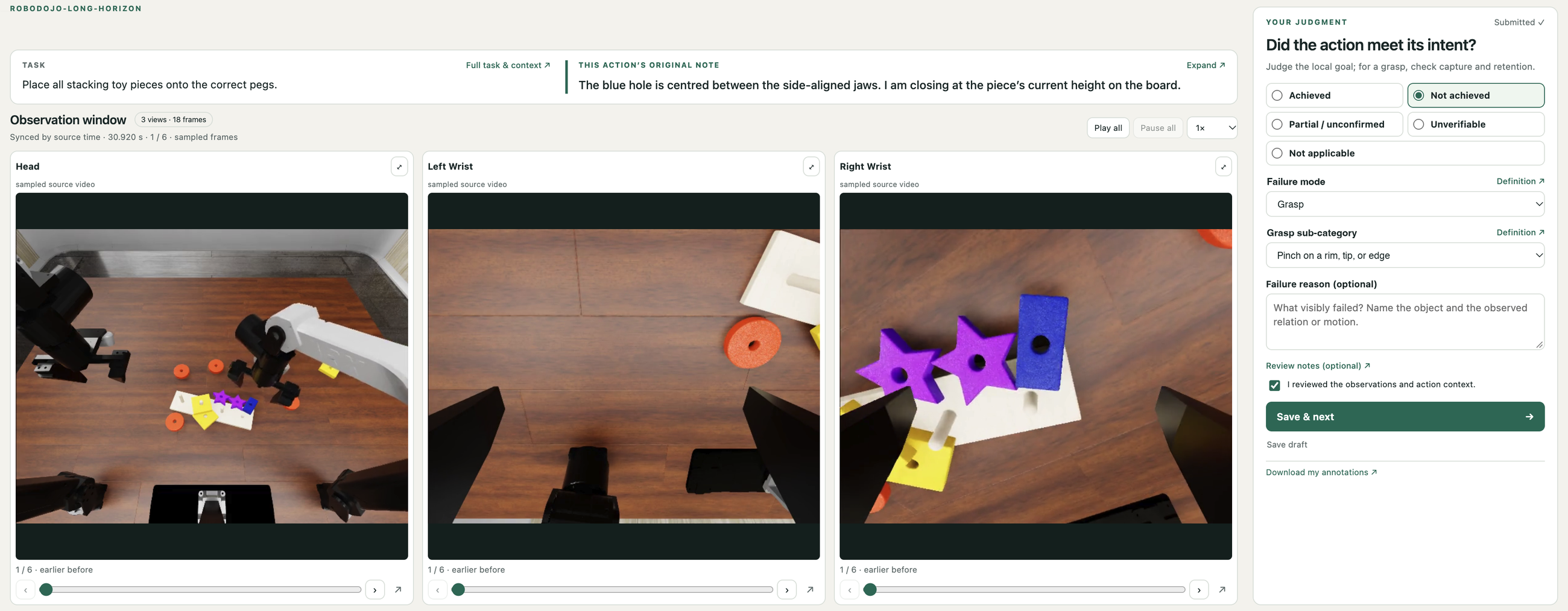}
\caption{\textbf{Annotation interface.} The annotator judges one action at a time from the task instruction, the local intent and synchronized frames of all camera views.}
\label{fig:supp_interface}
\end{figure*}

\cref{fig:supp_interface} shows the web-based interface used to label the human-annotated test set.
For each action, it shows the task instruction, the local intent stated with the action and the frames of every camera view, synchronized by source time.
The annotator selects one of the five outcomes and, for a failed action, its failure mode and, for \Grasp, its sub-category, with the definitions of \cref{sec:supp_taxonomy} linked next to each field.

\section{Critic Training Details}
\label{sec:supp_training}

\paragraph{Training data.}
The critic dataset draws on 2,027 episodes.
We keep executed actions with an achieved or failed outcome and complete pre- and post-action observations, and sample one achieved action for every three failed actions.
The splits are separated at the episode level, and no episode contributes actions to more than one split (\cref{tab:supp_critic_data}).
The 400 human-annotated actions used in \cref{tab:critic_eval} are drawn from the test split.

\begin{table}[!htbp]
\centering
\small
\setlength{\tabcolsep}{6pt}
\caption{\textbf{Splits of the critic dataset} in annotated actions.}
\label{tab:supp_critic_data}
\begin{tabular}{@{}lrrr@{}}
\toprule
Split & Achieved & Failed & Total \\
\midrule
Training & 3{,}211 & 9{,}633 & 12{,}844 \\
Validation & 387 & 1{,}161 & 1{,}548 \\
Test & 484 & 1{,}452 & 1{,}936 \\
\bottomrule
\end{tabular}
\end{table}

\paragraph{Input and target.}
Each sample follows the critic formulation of \cref{sec:critic_method}.
The input contains the task goal $G$, the local intent $g_t$ and the ordered camera frames of $\mathcal{O}_t$, preceded by a system prompt with the complete definitions of the outcome and failure-mode categories.
The target is a JSON object with the outcome, the failure mode, the grasp sub-category (\cref{tab:supp_subtypes}) and a short failure reason.
The last three fields are null for an achieved action.

\paragraph{Fine-tuning.}
We fine-tune Muse-Glimmer-30B with a LoRA adapter on the Fireworks AI platform, keeping the base weights frozen.
\cref{tab:supp_finetune} lists the key hyperparameters.

\begin{table}[htbp]
\centering
\caption{\textbf{Key hyperparameters} for fine-tuning Muse-Glimmer-30B as the critic.}
\label{tab:supp_finetune}
\small
\begin{tabular}{l c}
\toprule
\textbf{Configuration} & \textbf{Supervised Fine-tuning} \\
\midrule
Base model & Muse-Glimmer-30B \\
Trainable parameters & LoRA adapter only \\
LoRA rank & 32 \\
\midrule
Peak learning rate & $5 \times 10^{-5}$ \\
Learning rate schedule & Cosine with warmup \\
Warmup steps & 25 \\
\midrule
Training epochs & 2 \\
Batch size & 32 \\
Maximum context length & 65{,}536 \\
\bottomrule
\end{tabular}
\end{table}

\paragraph{Inference.}
The fine-tuned critic decodes at temperature 0 with at most 1{,}024 output tokens.
An output that is not valid JSON counts as incorrect.

\section{Qualitative Examples}
\label{sec:supp_examples}

\begin{figure*}[t]
\centering
\includegraphics[width=\textwidth]{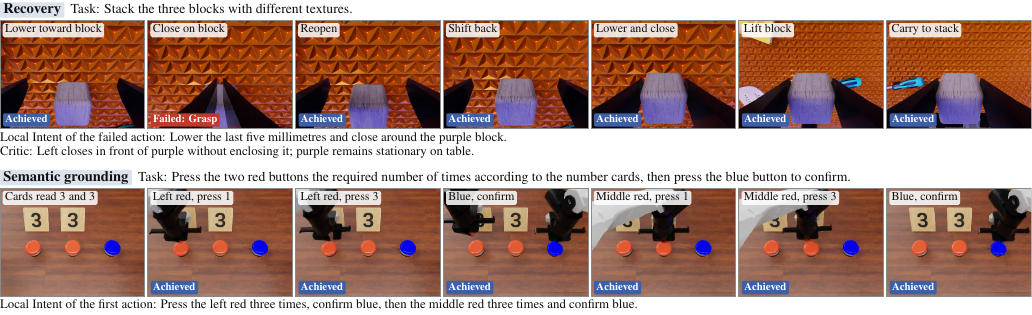}
\caption{\textbf{Closed-loop recovery and semantic grounding in two successful episodes.} Top, seven consecutive actions in which a missed grasp is corrected by a regrasp. Bottom, selected actions in which the agent presses the buttons in the count and order of its first local intent. Each frame shows the end of the action and the critic's outcome.}
\label{fig:supp_timelines}
\end{figure*}

\begin{figure*}[t]
\centering
\includegraphics[width=\textwidth]{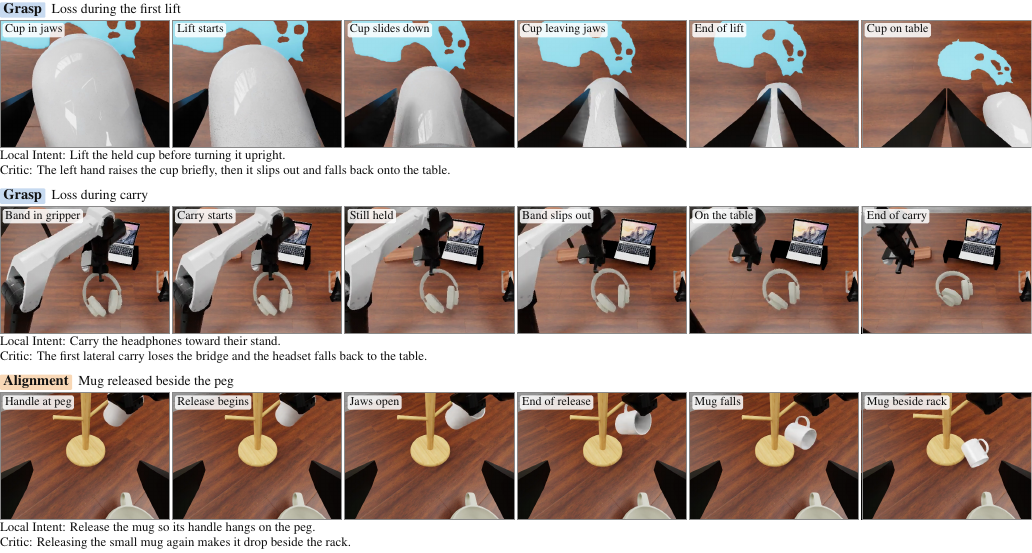}
\caption{\textbf{Failed actions over time.} Frames run from before the action to its end or shortly after. Each row gives the local intent and the critic's failure reason.}
\label{fig:supp_filmstrips}
\end{figure*}

\cref{fig:supp_timelines,fig:supp_filmstrips} show annotated RoboDojo actions with full-width frames.
\cref{fig:supp_timelines} follows two successful episodes, with one frame per action at the end of its execution and the critic's outcome.
In the first, the agent closes in front of the block, reopens, shifts back and closes on it, and then lifts and carries it, which illustrates closed-loop recovery (\cref{sec:exp_capabilities}).
In the second, every pressed button matches the count and order stated in the first local intent, which demonstrates semantic grounding (\cref{sec:exp_capabilities}).
\cref{fig:supp_filmstrips} follows single failed actions over time, together with the critic's failure reason.
The two \Grasp failures are losses after capture, during the first lift and during carrying (\cref{sec:grasp_analysis}), and the \Alignment failure releases a mug whose handle misses the peg.

\section{Additional Results on Direct Robot Control}
\label{sec:supp_additional}

We report additional experiments with GPT-6 Astra as the policy.
They cover policy refinement with critic feedback (\cref{sec:supp_refinement}), the effect of local intent on task success (\cref{sec:supp_intent_effect}), the zero-shot comparison on more tasks (\cref{sec:supp_zeroshot}) and inference latency (\cref{sec:supp_latency}).

\subsection{Action-Level Critic Guides Policy Refinement}
\label{sec:supp_refinement}

\begin{figure*}[t]
\centering
\includegraphics[width=\linewidth]{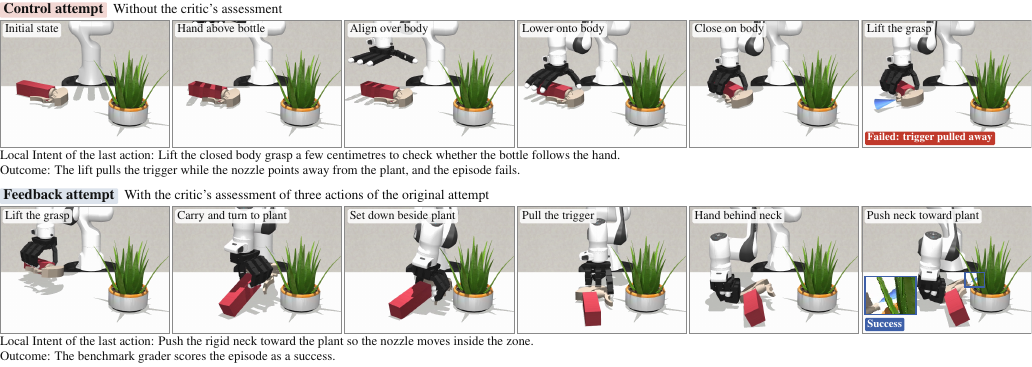}
\caption{\textbf{Control and feedback attempts on the DexJoCo water-plant task.} Each frame shows the scene at the end of the labeled action. The feedback row shows the first carry and the last actions of the episode, and the inset magnifies the spray in its last frame.}
\label{fig:supp_refinement_frames}
\end{figure*}

We investigate whether the action-level critique of a failed episode can guide the same policy on a new attempt.
On the Standard instance i00 of the DexJoCo water-plant task, the original GPT-6 Astra episode timed out without success.
We run two new attempts with the same frozen policy at high reasoning effort, the same seed and the same initial state.
Both attempts receive the task instruction, a notice that the previous attempt failed and three local intents from that attempt.
The feedback attempt also receives the critic's assessment of these three actions (\cref{fig:supp_feedback}), and the control attempt does not.
\cref{tab:supp_refinement} lists the three runs, and \cref{fig:supp_refinement_frames} shows frames of the two new attempts.
The control attempt pulls the trigger while the bottle points away from the plant and fails, and the feedback attempt completes the task.
The first critique reports that the nozzle still faced away from the plant after the turn (\Alignment), and the last local intent of the successful attempt is to push the rigid neck toward the plant so that the nozzle comes into range.
This case shows that action-level critiques can be returned to a frontier agent as feedback for its next attempt.

\begin{figure*}[t]
\centering
{\small
\setlength{\tabcolsep}{8pt}
\captionof{table}{\textbf{Repeated attempts on the DexJoCo water-plant task}, with the native score, the terminal outcome, the simulator calls and the control steps of each run.}
\label{tab:supp_refinement}
\begin{tabular}{@{}lclrr@{}}
\toprule
Run & Native score & Terminal outcome & Simulator calls & Control steps \\
\midrule
Original attempt & 0 & Policy timeout & 68 & 10{,}150 \\
Control attempt & 0 & Trigger pulled away from the plant & 12 & 575 \\
Feedback attempt & 1 & Success & 164 & 12{,}536 \\
\bottomrule
\end{tabular}
}
\vspace{8pt}
\begin{promptbox}{Context added to the task instruction in the feedback attempt}
\setlength{\parindent}{0pt}\setlength{\parskip}{0pt}
\textbf{Previous attempt}\par

The previous independent attempt on this same instance ended with native task success = false. You are starting a fresh instance. Use your own observations to decide what to do.\par
Selected local intent notes from the previous attempt, in execution order:\par
1. Carry the held bottle toward the plant and yaw the hand so the nozzle side turns toward the plant, without further squeezing.\par
2. Position the extended index and ring fingers above opposite sides of the fallen bottle body, with the middle finger and thumb tucked clear.\par
3. Lift and withdraw the head pinch from the pedestal, attempting to bring the bottle back above table height.\par

Historical Astra feedback on the previous attempt. These observations can be uncertain and describe the previous trajectory.\par
Prior note 1 --- M2: The bottle moves closer to the plant, but the trigger/nozzle side still faces away from it after the yaw, rather than toward the plant.\par
Prior note 2 --- M5: The recovery hand reaches over the bottle, but the thumb remains protruding beside the folded fingers rather than tucked clear as requested; the intended cleared approach is incomplete.\par
Prior note 3 --- M1 / S12: The withdrawing hand rises above table height without the spray bottle; the bottle remains on the pedestal foot below. The head pinch has not carried it out of the recovery location.\par
\end{promptbox}
\caption{\textbf{Feedback context of the repeated attempt.} The control attempt receives the same text without the last four lines, which hold the critic's assessment of the three actions.}
\label{fig:supp_feedback}
\end{figure*}

\subsection{Effect of Local Intent on Task Success}
\label{sec:supp_intent_effect}

Our policy rollouts require a local intent alongside every action (\cref{sec:critic_method}).
We investigate whether this requirement changes task performance by comparing our GPT-6 Astra episodes with the official GPT-6 Astra runs published by the Embodied First Exam, which do not require a local intent, on the 76 Standard instances that both cover.
GPT-6 Astra succeeds on 42 of the 76 instances (55.3\%) both with and without the local intent requirement.
This shows that stating a local intent alongside every action has little effect on the success rate of GPT-6 Astra, and that the local intents that AgentRoboCritic relies on come at little cost to policy performance.

\subsection{Zero-Shot Comparison on More Tasks}
\label{sec:supp_zeroshot}

\begin{figure}[t]
\centering
\includegraphics[width=0.6\linewidth]{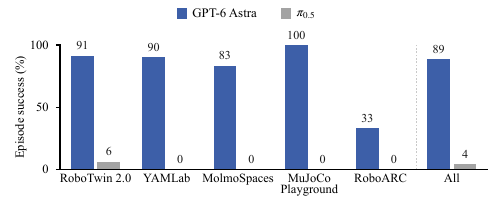}
\caption{\textbf{Zero-shot success rates on matched tasks}, the percentage of tasks in each suite that each policy completes. All combines the five suites.}
\label{fig:supp_zeroshot}
\end{figure}

We extend the zero-shot comparison of \cref{fig:zeroshot_comparison} to matched tasks from five suites of the Embodied First Exam, on which both GPT-6 Astra and $\pi_{0.5}$ have a scored episode.
Both policies run each task instance without task-specific fine-tuning or demonstrations, and the grader of each benchmark scores the episode.
$\pi_{0.5}$ sends a low-level action to the simulator after every observation, and GPT-6 Astra issues numerical actions through tool calls (\cref{sec:direct_control}).
\cref{fig:supp_zeroshot} shows the success rates of both policies.
GPT-6 Astra outperforms $\pi_{0.5}$ on every suite, and every task that $\pi_{0.5}$ completes is also completed by GPT-6 Astra.
This further supports the zero-shot advantage of frontier agents shown in \cref{fig:zeroshot_comparison}.

\begin{table}[t]
\centering
\small
\setlength{\tabcolsep}{6pt}
\caption{\textbf{Decision latency}, the median over tasks of each run's median interval from an observation to the next executed action, in seconds, with the interquartile range in brackets.}
\label{tab:supp_latency}
\begin{tabular}{@{}lcc@{}}
\toprule
Tasks & GPT-6 Astra (s) & $\pi_{0.5}$ (s) \\
\midrule
RoboTwin~2.0 & 38.4 [19.2, 54.7] & 0.109 [0.108, 0.111] \\
YAMLab & 34.1 [15.5, 45.4] & 0.142 [0.141, 0.143] \\
All matched tasks & 31.9 [4.8, 46.8] & 0.110 [0.108, 0.128] \\
\bottomrule
\end{tabular}
\end{table}

\subsection{Inference Latency}
\label{sec:supp_latency}

We measure the decision latency of GPT-6 Astra and $\pi_{0.5}$ on the tasks of the zero-shot comparison (\cref{fig:zeroshot_comparison}) and on further matched tasks of the Embodied First Exam suites.
A latency interval runs from the completion of the most recent observation request to the start of the next executed action submission, both read from the journal of the robot service.
Read-only tool calls inside the interval are included, and intervals across resets or episodes and actions that are not executed are excluded.
For each run we take the median interval, and \cref{tab:supp_latency} reports the median and interquartile range of these run medians over the tasks timed for both policies.
Across 74 matched tasks, the median decision takes 31.9\,s for GPT-6 Astra and 0.11\,s for $\pi_{0.5}$, and the gap is similar on RoboTwin~2.0 and YAMLab.

\section{Verification Against Critic Self-Preference Bias}
\label{sec:supp_bias}

GPT-6 Astra serves as both the control policy and the primary critic in our study.
We therefore examine whether using the same model introduces self-preference bias, and whether fine-tuning on our dataset transfers such bias to the open-source critic.

\paragraph{Frontier critics.}
A critic biased toward its own policy would systematically label failed actions as \Achieved.
\cref{tab:supp_leniency} counts these errors for each critic on the human-annotated test set.
GPT-6 Astra labels only 19 of the 264 human-labeled failed actions as \Achieved, the fewest among all evaluated critics.
Claude Fable 5.2, a frontier model from a different model family, reaches a similar outcome accuracy of 78.25\%, compared to 79.25\% for GPT-6 Astra (\cref{tab:critic_eval}).
This shows that GPT-6 Astra is the strictest evaluator of its own failed actions, and that the high agreement with human labels holds across model families.

\begin{table}[t]
\centering
\small
\setlength{\tabcolsep}{6pt}
\caption{\textbf{Failed actions labeled \Achieved} by each critic, out of the 264 actions that the annotators judge as \Failed in the human-annotated test set.}
\label{tab:supp_leniency}
\begin{tabular}{@{}lr@{}}
\toprule
Critic & Labeled \Achieved \\
\midrule
GPT-6 Astra & 19 (7.2\%) \\
Claude Fable 5.2 & 75 (28.4\%) \\
Claude Opus 5.5 & 77 (29.2\%) \\
GPT-6.1 Sol & 81 (30.7\%) \\
GPT-6.1 Terra & 71 (26.9\%) \\
\midrule
Muse-Glimmer-30B (Zero-Shot) & 67 (25.4\%) \\
Muse-Glimmer-30B (Fine-Tuned) & 34 (12.9\%) \\
\bottomrule
\end{tabular}
\end{table}

\paragraph{Fine-tuned critic.}
Muse-Glimmer-30B is fine-tuned on our dataset of actions annotated by GPT-6 Astra and scored against the human labels, on test episodes that are kept out of training (\cref{sec:critic_reliability,sec:supp_training}).
Fine-tuning halves the failed actions that it labels as \Achieved (\cref{tab:supp_leniency}) and raises its failure-mode accuracy above every critic except GPT-6 Astra (\cref{tab:critic_eval}).
This shows that fine-tuning on our dataset makes the critic stricter on failed actions and more accurate on the human-annotated benchmark.

\section{Critic Codebook}
\label{sec:supp_codebook}

The codebook below follows the system prompt of \cref{fig:supp_prompt} in every critic request.
It gives a label, a definition, an explanation, a boundary rule and an example for each failure mode, and a label and a definition for each \Grasp sub-category.
\cref{tab:supp_codes} lists the names used in this paper for its identifiers.

\begin{codebookbox}{Codebook}
\setlength{\parindent}{0pt}\setlength{\parskip}{4pt}
\textbf{M1\enspace Acquisition / retention.} \textit{Definition.} The required object is not captured, is lost, or is already absent from the manipulating hand. \textit{Explanation.} Use when the current local manipulation fails because the required object is not captured, slips out, is dislodged, or was already absent from the hand before the call. Identify the object, arm, contact geometry, and stage of loss to the extent supported by the images. \textit{Boundary.} Add an S1-S14 subcategory. Distinguish a new loss from an already-empty follow-up (S11), and retain S12 when non-retention is visible but the contact error is not. A deliberate release or a successful diagnostic test is not automatically an M1 failure. An object still held but offset from its target is usually M2. \textit{Example.} The note asks to grasp the coin, but the jaws close above its top and the subsequent lift leaves it on the table: M1 with S1.\par
\textbf{M2\enspace Targeting / engagement.} \textit{Definition.} A held object or commanded contact remains offset, tilted, or outside its intended target. \textit{Explanation.} Use when a held object or intended contact misses the current local target because it remains offset, misoriented, too shallow, or outside the intended contact region. Describe the visible target relation rather than attributing the error to an unobserved controller or perception defect. \textit{Boundary.} The immediate problem is targeting or engagement, not missing possession (M1). A major unintended excursion or collision can be M3. If contact is established but the intended tool effect is absent, consider M4. A small intermediate approach is not a failed insertion unless insertion was the current stated goal. \textit{Example.} The note requests inserting the retained plug, but the plug tip stays beside the socket opening at the end of the call.\par
\textbf{M3\enspace Motion / interference.} \textit{Definition.} The current call causes a major unintended excursion, collision, or object displacement. \textit{Explanation.} Use when the current call produces a major unintended motion, collision, or displacement that defeats its local goal. Record which hand or object moved, what it contacted, and the unwanted change; do not label every incidental touch a collision failure. \textit{Boundary.} An ordinary positional miss can be M2. When closure pushes the intended object away and thereby prevents capture, the more specific acquisition description is M1/S4. A later call should not inherit M3 solely because an earlier call disturbed the scene. \textit{Example.} During a requested approach, the forearm strikes a neighboring stack and knocks several blocks off their supports.\par
\textbf{M4\enspace Task-effect failure.} \textit{Definition.} A retained tool/object fails to produce its intended pour, strike, or other task effect. \textit{Explanation.} Use when the relevant object or tool remains retained but the call does not produce its stated physical effect, such as a pour, strike, scrape, or target-state change. Identify the absent effect and the visible interaction that supports the judgment. \textit{Boundary.} If the hand is empty, use M1. If the tool visibly misses its contact target, M2 may be the more specific mechanism. If the requested movement or opening itself never completes, consider M5. Do not infer an absent effect that the released observations cannot show. \textit{Example.} The hand retains the container and executes the intended tilt, but no contents enter the receiving bowl despite the note requesting a pour into it.\par
\textbf{M5\enspace Incomplete motion / reset.} \textit{Definition.} A movement, opening, withdrawal, or return is incomplete without a more specific mechanism. \textit{Explanation.} Use when a requested movement, opening, withdrawal, reset, or return remains incomplete and no more specific acquisition, targeting, interference, or task-effect mechanism is supported. Describe the commanded change and the visible residual state. \textit{Boundary.} This code requires a demonstrably unachieved local goal; an intentionally incremental movement or a truncated, unobservable result may instead be partial/unconfirmed or unverifiable. Rejected commands are not executed-action M5 failures. Do not infer why the mechanism failed to move. \textit{Example.} The note requests opening the gripper and releasing the object, but the fingers remain closed and the object remains clamped at the supported endpoint.\par
\textbf{M6\enspace Unresolved mechanism.} \textit{Definition.} The local goal fails, but the available evidence does not identify a more specific mechanism. \textit{Explanation.} Use when the current local goal is demonstrably not achieved but the available evidence does not identify a more specific action mechanism. State the failed local result, identify the missing or occluded evidence, and avoid choosing a visually unsupported mechanical explanation. \textit{Boundary.} M6 is different from an unverifiable outcome: failure itself must be supported. If non-retention is established but its contact geometry is unclear, use M1/S12 rather than M6. A missing original note is not by itself evidence of a failed local goal. \textit{Example.} The visible local target state is unchanged after an intended manipulation, but the relevant interaction is occluded and the evidence cannot distinguish incomplete motion from a failed task effect.\par
\textbf{M7\enspace Wrong local choice.} \textit{Definition.} The stated local selection or relation itself is unsuccessful; a separate task mismatch is not sufficient. \textit{Explanation.} Use when the call fails its own stated selection, ordering, or categorical-relation goal. The identifying issue is which object, answer, or relation was enacted, even if the low-level movement completed normally. The original local note and the relevant reference must support that discrepancy. \textit{Boundary.} A mismatch with the overall task is not sufficient if the action faithfully executes its own note; this does not by itself establish a failed local action. Distinguish choosing the wrong named target from a continuous targeting miss (M2). M7 remains unassessed in the preserved RoboDojo-Precision baseline. \textit{Example.} The note requests pressing the key labeled 3, but the robot presses the clearly labeled 7 key. If the note itself requests 7 and the robot presses 7, that is not M7 merely because the overall task required 3.\par
\textbf{S1\enspace Jaws close above object.} Fingers close over the top rather than around the body; a vertical miss must be visible.\par
\textbf{S2\enspace Jaws close beside / ahead / behind.} The object is outside the opposing jaw gap; record the visible lateral or depth offset.\par
\textbf{S3\enspace Only rim / tip / edge pinched.} Only a shallow rim, tip, edge, or tapered surface is caught; name the contact region.\par
\textbf{S4\enspace Closure pushes / tips object away.} The acquiring hand pushes, rolls, or tips the object instead of enclosing it.\par
\textbf{S5\enspace Support / another object caught.} A rest, fixture, support, or other object is caught instead of the intended target.\par
\textbf{S6\enspace Receiving hand misses transfer.} The receiving jaws fail to secure the transferred object; donor timing is not assumed.\par
\textbf{S7\enspace Slip during initial lift.} After transient capture, the object separates during the first lift; it did leave its support.\par
\textbf{S8\enspace Slip during carry / wrist turn.} An object that previously moved with the hand separates during carrying or turning.\par
\textbf{S9\enspace Contact pulls object out.} Visible contact with a surface, target, obstacle, or other arm dislodges a held object.\par
\textbf{S10\enspace Opening before support.} The gripper opens before a secure support is established and the object falls.\par
\textbf{S11\enspace Already-empty follow-up motion.} The hand was already empty before this call; it is a downstream motion, not a new loss.\par
\textbf{S12\enspace Contact error not visible.} Non-retention is established, but offset, shallow contact, and transient slip cannot be resolved.\par
\textbf{S13\enspace Wrong jaw faces / axis.} The jaw orientation visibly fails to oppose the faces or axis that must be pinched.\par
\textbf{S14\enspace Inside-hole nut grasp fails.} Visible contact inside a nut bore fails to lift the nut; internal contact must be observed.\par
\end{codebookbox}

\medskip
\noindent\begin{minipage}{\linewidth}
\centering
\setlength{\tabcolsep}{4pt}
\captionof{table}{\textbf{Codebook identifiers}, with their codebook labels and the shorter names used in the paper.}
\label{tab:supp_codes}
\small
\begin{tabular}{@{}lll@{}}
\toprule
Identifier & Codebook label & Short name in the paper \\
\midrule
\multicolumn{3}{@{}l}{\textit{Failure modes}} \\
M1 & Acquisition / retention & \Grasp \\
M2 & Targeting / engagement & \Alignment \\
M3 & Motion / interference & \Interference \\
M4 & Task-effect failure & \Effect \\
M5 & Incomplete motion / reset & \IncMotion \\
M6 & Unresolved mechanism & \Unresolved \\
M7 & Wrong local choice & \Choice \\
\multicolumn{3}{@{}l}{\textit{Grasp sub-categories}} \\
S1 & Jaws close above object & Closure above the object \\
S2 & Jaws close beside / ahead / behind & Closure beside the object \\
S3 & Only rim / tip / edge pinched & Pinch on a rim, tip, or edge \\
S4 & Closure pushes / tips object away & Closure pushes or tips the object away \\
S5 & Support / another object caught & The support or another object is caught \\
S6 & Receiving hand misses transfer & Receiving hand fails to secure a transfer \\
S7 & Slip during initial lift & Loss during the first lift \\
S8 & Slip during carry / wrist turn & Loss during carry or wrist turning \\
S9 & Contact pulls object out & External contact dislodges the object \\
S10 & Opening before support & Opening before support is established \\
S11 & Already-empty follow-up motion & Action with an empty gripper \\
S12 & Contact error not visible & Unresolved contact \\
S13 & Wrong jaw faces / axis & Wrong jaw faces or axis oppose the contact \\
S14 & Inside-hole nut grasp fails & Failed nut-bore grasp \\
\bottomrule
\end{tabular}
\end{minipage}
\medskip

\end{document}